\documentclass[journal]{IEEEtran}
\usepackage{bbm}
\usepackage{graphicx}
\usepackage{amsmath,amssymb} 
\usepackage{color}
 \usepackage{slashbox}
 \usepackage[tight,footnotesize]{subfigure}
 \usepackage{footnote}
 \usepackage{tablefootnote}
\usepackage{dsfont}
\usepackage{mathtools} 
\usepackage{mathrsfs}
\usepackage{url}
\usepackage{hyperref}

\ifCLASSINFOpdf
\else
\fi
\begin{document}
%
\title{Forward Stagewise Additive Model for Collaborative Multiview Boosting}
%
%
%

\author{Avisek~Lahiri, Biswajit Paria,
        Prabir~Kumar~Biswas, \textit{Senior Member, IEEE}
\thanks{A.Lahiri, and P.K.Biswas are with the Dept. of E\&ECE and B.Paria is with Dept. of CSE, Indian Institute of Technology-Kharagpur,
West Bengal-721302, India. E-mail: avisek@ece.iitkgp.ernet.in}
}

\maketitle

\begin{abstract}
Multiview assisted learning has gained significant attention in recent years in supervised learning genre. Availability of high performance computing devices enables learning algorithms to search simultaneously over multiple views or feature spaces to obtain an optimum classification performance. The paper is a pioneering attempt of formulating a mathematical foundation for realizing a multiview aided collaborative boosting architecture for multiclass classification. Most of the present algorithms apply multiview learning heuristically without exploring the fundamental mathematical changes imposed on traditional boosting. Also, most of the algorithms are restricted to two class or view setting. Our proposed mathematical framework enables collaborative boosting across any finite dimensional view spaces for multiclass learning. The boosting framework is based on forward stagewise additive model which minimizes a novel exponential loss function. We show that the exponential loss function essentially captures difficulty of a training sample space instead of the traditional `1/0' loss. The new algorithm restricts a weak view from over learning and thereby preventing overfitting. The model is inspired by our earlier attempt \cite{accv} on collaborative boosting which was devoid of mathematical justification. The proposed algorithm is shown to converge much nearer to global minimum in the exponential loss space and thus supersedes our previous algorithm. The paper also presents analytical and numerical analysis of convergence and margin bounds for multiview boosting algorithms and we show that our proposed ensemble learning manifests lower error bound and higher margin compared to our previous model. Also, the proposed model is compared with traditional boosting and recent multiview boosting algorithms. In majority instances the new algorithm manifests faster rate of convergence on training set error and simultaneously also offers better generalization performance. Kappa-error diagram analysis reveals the robustness of the proposed boosting framework to labeling noise.
\end{abstract}


%
\IEEEpeerreviewmaketitle

\section{Introduction}

\IEEEPARstart{M}{ultiview}supervised learning has achieved significant attention among machine learning practitioners in recent times. In today's Big Data platform it is quite common that a single learning objective is represented over multiple feature spaces.  To appreciate this, let us consider the KDD Network Intrusion Challenge \cite{kdd99}. In this challenge, domain experts identified four major variants of network intrusion and characterized them over three feature spaces, viz. TCP components, content features and traffic features. Another motivating example is the `100 Leaves Dataset' \cite{leaves}, where the objective is to classify hundred classes of leaves. Each leaf is characterized by shape, margin and texture features. Multiview representation of objective function is also common in other disciplines such as drug discovery \cite{drug1}, medical image processing \cite{spie_multiview}, dialogue classification \cite{mumbo_speech}, etc.
\par One intuitive method is to combine all the features and then train a classifier on a reduced dimensional feature space. But dimensionality reduction has its own demerits. Usually features are engineered by experts and each feature has its own physical significance. Projecting the features onto a reduced dimensional space usually obscures the physical interpretation of the reduced feature space. Another problem with dimensionality reduction is that the subtle features are lost during the projection process. These features have been shown to foster better discriminative capability in presence of noisy data \cite{isi1, dimension_survey}. Training by the above method is sometimes referred to as Early Fusion. Another paradigm of multiview learning is Late Fusion; the objective is to separately learn classifiers on each feature space and finally conglomerate the classifiers by majority voting \cite{late_fusion}. The major issue is that these algorithms do not incorporate collaborative learning across views. We feel that it is an interesting strategy to communicate classification performance over views and model weight distribution over sample space  according to this communication. Also, performance of fusion techniques are problem specific and thus the optimum fusion strategy is unknown a priori \cite{late_vs_early}. 
\par Multiview learning has been an established genre of research in semi supervised learning where manual annotation labor is reduced by stochastically learning over labeled and unlabeled training examples. Query-by-committee \cite{qbc} and co-training \cite{cotrain} were the two pioneering efforts in this direction. For these algorithms, the objective function is represented over two mutually independent and sufficient view spaces. Independent classifiers are trained on each view space using the small number of labeled examples. The remaining unlabeled instance space is annotated by iterative majority voting of the classifiers trained on the two views. Recently, Co-training by committee   \cite{cotraincommittee} obviates the constraint of mutual orthogonality of the views. Significant success of multiview learning in semi supervised learning has been the primary motivation of our work.
\par The paper presents the following notable contributions:
\begin{enumerate}
	\item To the best of our knowledge this is the pioneering attempt in formulating a additive model based mathematical framework for multiview collaborative boosting. It is to be noted that the primary significance of our current work is to mathematically bolster our previous attempt of multiview learning, MA-AdaBoost \cite{accv}, which was based on intuitive cues.
	\item Stagewise modeling of boosting requires a loss function and in this regard we propose a novel exponential multiview weighted loss function to grade strata of `difficultiness' of an example. Using this loss function, we were able to derive a similar multiview weight update criterion used in \cite{accv}; this signifies the aptness of our present analytical approach and the correctness of our previous intuitive modeling.
	\item  We devise a two step optimization framework for converging much nearer to global minimum of the proposed exponential loss space compared to our previous attempt of MA-AdaBoost
	\item Analytical expressions are derived for upper bounding training set error and margin distribution under multiview boosting setting. We numerically study the variations of these bounds and show that the proposed framework is superior compared to MA-AdaBoost
	\item Extensive simulations are performed on challenging datasets such as 100-Leaves \cite{leaves}, Eye classification \cite{icvgip}, MNIST hand written character recognition and 11 different real world datasets from UCI database \cite{uci}. We compare our model with traditional and state-of-the-art multiclass boosting algorithms
	\item Kappa-Error visualization is studied to manifest robustness of proposed SAMA-AdaBoost to labeling noise.
\end{enumerate}

\par The rest of the paper is organized as follows. Section II gives a brief overview of traditional and variants of AdaBoost. Section III presents some recent works on multiview boosting algorithms and how our work addresses some of the short comings of existing algorithms. Section IV formally describes our collaborative boosting framework followed by convergence and margin analysis in Section V. Experimental analysis are presented in Section VI. Finally, we conclude the paper in Section VII concludes the paper with a brief discussion and future extensions of the proposed work.


\section{Brief Overview on Adaptive Boosting}
In this section we present a brief overview of the traditional adaptive boosting algorithm \cite{boost} and the recent variants of AdaBoost. Also, we discuss some of the mathematical viewpoints which bolster the principle of AdaBoost.
Suppose we have been provided with a training set $X=\{$$(x_1,l_1)$, $(x_2,l_2)$....$(x_n,l_n)$$\}$, where $x_i \in R^d$  denotes $d$-dimensional input variable and $l_i \in \{1,2,...L\}$ is the class label. The fundamental concept of AdaBoost is to formulate a weak classifier in each round of boosting and ultimately conglomerate the weak classifiers into a superior meta-classifier. AdaBoost initially maintains an uniform weight distribution over training set and builds a weak classifier. For the next boosting round, weights of misclassified examples are enhanced while weights of correctly classified examples are reduced. Such a modified weight distribution aids the next weak classifier to focus more on misclassified examples and the process continues iteratively. The final classifier is formed by linear weighted combination of the weak classifiers. AdaBoost.MH \cite{adaboost_mh} is usually used for multiclass classification using `one-versus-all' strategy.
\par Traditional AdaBoost has undergone plethora of modifications due to active interest among machine learning community. WNS-Boost \cite{wns} uses a weighted novelty sampling algorithm to extract the most discriminative subset from the training sample set. The algorithm then runs AdaBoost on the reduced sample space and thereby enhances speed of training with minimal loss of accuracy. SampleBoost \cite{sample_boost} is aimed to handle early termination of multiclass AdaBoost and to destabilize weak learners which repeatedly misclassify same set of training examples. Zhang et al. \cite{prob_adaboost} introduces a correction factor for reweighting scheme of traditional AdaBoost for enhanced generalization accuracy. 
\par Researchers have used margin analysis theory \cite{margin_boost1,margin_boost2} to explain working principle of AdaBoost. Another view point of explaining AdaBoost is functional gradient descent \cite{function_boost1}. A modish way of explaining AdaBoost is forward stagewise additive model which minimizes an exponential loss function \cite{discuss_boost}. Inspired by the model in \cite{discuss_boost}, Zhu et al. proposed SAMME \cite{samme} for multiclass boosting using Fisher-consistent exponential loss function.
\par We wish to acknowledge that \cite{discuss_boost,samme} have been instrumental in our thought process for the proposed 
algorithm but we differ on several aspects. As per our best knowledge this is the first attempt to formulate a mathematical model for multiview boosting using stagewise modeling. Also, the existing mathematical frameworks which explain boosting lack the scope of scalable collaborative learning.
\section{Related Works on Multiview Boosting}
 The current work is motivated by our previous successful attempt on multiview assisted adaptive boosting, MA-AdaBoost \cite{accv}. MA-AdaBoost is the first attempt to grade the difficulty of a training example instead of the traditional `1/0' loss usually practised in boosting genre.  We have successfully used MA-AdaBoost in computer vision applications\cite{icvgip} and other real world datasets. 
But MA-AdaBoost is primarily based on heuristics. The objective of this paper is to understand and enhance the performance of MA-AdaBoost by formulating a thorough mathematical justification.
\par Recently, researchers have proposed different algorithms for group based learning. 2-Boost  \cite{boost2} and Co-AdaBoost \cite{coadaboost} are closely related to each other. Both of these algorithms maintain a single weight distribution over the feature spaces and weight update depends on ensemble performance. Our algorithm is considerably different from these two algorithms. The proposed algorithm is scalable to any finite dimensional view and label space while 2-boost and Co-AdaBoost is restricted to two class and view setting. 2-Boost additionally requires that the two views be learnt by different baseline algorithms. Moreover, these two algorithms formulate the final hypothesis by majority voting. In contrast, our model uses a novel scheme of reward-penalty based voting. Share-Boost \cite{shareboost} has got some similarities with 2-Boost except that after each round of boosting Share-Boost discards all weak learners except the globally optimum learner (classifier with least weighted error).
         \par AdaBoost.Group \cite{adaboost.group} was proposed for group based learning in which the authors assumed that sample space can be categorized into discriminative groups. Boosting was performed in group level and independent classifiers were optimally trained by maximizing F-score on individual views. The final classifier was reported using majority voting over all the groups. Separately training independent classifiers inhibits AdaBoost.Group from inter-view collaboration. Also, optimizing classifiers over each local view space does not ensure to optimize the final global classifier.
     \par Mumbo is an elegant example of multiview assisted boosting algorithm \cite{mumbo1,mumbo2}. The fundamental idea of Mumbo is to remove an arduous example from view space of weak learners and simultaneously increase weight of that example in view space of strong learners. A variant of Mumbo has been used by Kwak et al. \cite{spie_multiview} for tissue segmentation. Mumbo maintains cost matrix $\mathbf{C_k}$ on each view space $k$, where $C_k(i,j)$ represents cost of classifying training example $x_i$  belonging to class $i$ to class $j$ on view $k$. The total space requirement for Mumbo is $\mathcal{O}(Q.n.L)$ where $ Q$ and $L$ denote total number of views and classes respectively while $n$ is number of training samples. Such a space requirement is debatable in case of large datasets. Our proposed algorithm is void of such space requirements. Moreover, Mumbo requires that atleast one view should be `strong' which is aided by other `weak' views. Selection of a strong view in case of large dataset is not a trivial task. Our proposed algorithm adaptively assigns importance to a view space during run time and so end users need not manually specify a strong view.
 \section{Collaborative Boosting Framework}
In this section we formally introduce our proposed framework for stagewise additive multiview assisted boosting algorithm, SAMA-AdaBoost. We consider the most general case where an example $x_i$ is represented over total $V$ views and the corresponding class label $y_i\in\{1,2,...K\}$.

\subsection{Formulation of Exponential Loss Function}
If an example $x_i$ belongs to class $c$, then we assign a corresponding label vector ${Y}_i=[0~0~....1~ 0~ 0 ....]^T$,  such that there are $K-1$ zeroes and the $c^{th}$ element of  ${Y}_i$, represented by $y_{i,c}=1$. We denote a weak hypothesis vector learnt on view space $j$ after $t$ boosting rounds as ${h_j^t}(x)$ and  $h_{j,i}^t(x)$ represents $i^{th}$  element of ${h_j^t}(x)$. Before we delve into formulation of the exponential loss function, we need to pre-process the hypothesis vectors. Specifically, the $k^{th}$ element of the hypothesis vector ${h_j^t}(x_i)$ on $x_i$ is modified by the following equation;
$${\widetilde{h_{j,k}^t(x_i)}}=$$$$\delta({Y_i^T}.{h_j^t}(x_i)).\{((-1)^{y_{i,k}-h_{j,k}^t(x_i)}).\delta(y_{i,k}+h_{j,k}^t(x_i)-1)\}$$
\begin{equation}
+~ h_{j,k}^t(x_i).\delta({Y_i^T}.{h_j^t}(x_i)-1)
\label{eq_transform}
\end{equation}
where $\delta(x)$=1 only for $x$=0 and zero elsewhere. The first part of Eq.(\ref{eq_transform}) triggers in case of misclassified vectors because in case of misclassification, ${Y_i^T}.{h_j^t}(x_i)$=0. The second $\delta(.)$ function in first part of Eq. \ref{eq_transform} is triggered only if the corresponding $k^{th}$ entry of either ${Y_i}$ or ${h_j^t(x_i)}$ is `1' and in those cases, power term transforms the elements $h_{j,k}^t(x_i)$ to `-1'.  The second part of Eq.(\ref{eq_transform}) triggers in case of correctly classified vector but keeps the vector intact. Table \ref{table_conversion} delineates a representation of the above transformation process where we consider an example, $x_i$, to belong to class 1.
\begin{table}
\caption{An illustration to explain the transformation of hypothesis vectors ${h_1^t}(x_i)$, ${h_2^t}(x_i)$. For illustration purpose we show example using only 2-views. ${h_1^t}(x_i)$ and ${h_2^t}(x_i)$ are correct and incorrect hypothesis vector respectively.}
\centering
\begin{tabular}{c c c c c}
\hline\hline\\
${Y_i}$ &${h_1^t}(x_i)$ & Transformed : ${\widetilde{h_1^t}}(x_i)$ & ${h_2^t}(x_i)$&\hspace{-3mm}Transformed : ${\widetilde{h_2^t}}(x_i)$\\
\hline\\
1&1&1&0&-1\\
0&0&0&0&0\\
0&0&0&1&-1\\
0&0&0&0&0\\
.&.&.&.&.\\
.&.&.&.&.\\
\hline
\end{tabular}
\label{table_conversion}
\end{table}
We define an exponential loss function $L({Y_i},\sum_{j=1}^V{\widetilde{h_j^t}(x_i)})$
$$L({Y_i},\sum_{j=1}^V{\widetilde{h_j^t}(x_i)})=exp\left( \frac{-\sum_{j=1}^V\sum_{k=1}^K y_{i,k}.\widetilde{h_{j,k}^t(x_i)} }{V}\right)$$
\begin{equation}
~~~~~~~~~~=exp\left( \frac{-\sum_{j=1}^V{Y_i}^T.{\widetilde{h_j^t}(x_i)}}{V} 
  \right)
  \label{eq_2}
\end{equation}
where $V$ is the total number of feature spaces or views. From Table \ref{table_conversion} we see that if ${{h_j^t}(x_i)}$ is a correct classification vector, then ${Y_i^T.\widetilde{h_j^t}(x_i)}=1$ else ${Y_i^T.\widetilde{h_j^t}(x_i)}=-1$. If $x_i$ is misclassified by weak learners on total $b_i$ views, then

\begin{equation}
L({Y_i},\sum_{j=1}^V{\widetilde{h_j^t}(x_i)})=exp\left(\frac{(V-b_i)-(b_i)}{-V} \right)
\end{equation}
\begin{equation}
~~~~~~~~~=exp\left(\frac{2 b_i}{V}-1 \right)
\label{eq_4}
\end{equation}
We argue that the term $\left(\frac{2 b_i}{V}-1 \right)$ in Eq.(\ref{eq_4}) manifests the difficulty of $x_i$. Weak classifiers over all views are trying to learn $x_i$. So, it makes sense to judge difficulty of $x_i$ in terms of total misclassified views and incorporate this graded difficulty in the loss function which will eventually govern the boosting network.


\subsection{Forward Stagewise Model for SAMA-AdaBoost}
In this section we present a forward stagewise additive model to understand the working principle of our proposed SAMA-AdaBoost. We opt for a greedy approach where in each step we optimize one more weak classifier and add it to existing ensemble space. Specifically, the approach can be viewed as stagewise learning of additive models \cite{discuss_boost} with initial ensemble space as null space. We define $f_k^M(x)$ as the additive model learnt over $M$ boosting rounds on a particular view space $k$:
\begin{equation}
f_k^M(x)=\sum_{t=1}^M \beta^t{\widetilde{h_k^t(x)}}
\end{equation}
where $\beta^t>0 \in \mathbb{R}$ denotes learning rate. Our goal is to learn the meta-model $F_V^M$ which represents the overall additive model learnt over $M$ boosting rounds on total $V$ views.
\begin{equation}
F_V^M=\sum_{k=1}^V f_k^M(x)
\end{equation}
So, after any arbitrary $m$ (boosting rounds) and $v$ (total number of views) we can write,
\begin{equation}
F_v^m=\sum_{k=1}^v\sum_{t=1}^m\beta^t {\widetilde{h_k^t(x)}}
\end{equation}
\begin{equation}
=\sum_{k=1}^v\sum_{t=1}^{m-1}\beta^t {\widetilde{h_k^t(x)}}+\sum_{k=1}^v\beta^m{\widetilde{h_k^m(x)}}
\end{equation}
\begin{equation}
=\sum_{k=1}^vf_k^{m-1}(x)+\beta^m\sum_{k=1}^v{\widetilde{h_k^m(x)}}
\label{eq_9}
\end{equation}
The first part of Eq.(\ref{eq_9}) i.e., $\sum_{k=1}^vf_k^{m-1}(x)$ represents part of our model which has already been learnt and hence we cannot modify it. Our aim is to optimize the second part of Eq.(\ref{eq_9}) i.e., $\beta^m\sum_{k=1}^v{\widetilde{h_k^m(x)}}$. Here, we will make use of our proposed exponential loss function as reported in Eq.(\ref{eq_2}). The solution for the next best set of weak classifiers and learning rate on $m^{th}$ boosting round can be written as:
$$\left(\beta^{m*}, {\sum_{k=1}^v\widetilde{h^{k*}(x_i)}} \right)=$$
\begin{equation}
\underset{\beta,{\sum_{k=1}^v\widetilde{h^k(x_i)}}} {\mathrm{argmin}}  \sum_{i=1}^nexp\left[\frac{-{Y_i}^T}{v} \left( \sum_{k=1}^vf_k^{m-1}(x_i)+\beta \sum_{k=1}^v {\widetilde{h^k(x_i)}}      \right) \right]
\end{equation}
\begin{equation}
=\underset{\beta,{\sum_{k=1}^v\widetilde{h^k(x_i)}}} {\mathrm{argmin}} \sum_{i=1}^nW_i(m)exp\left(\frac{-\beta{Y_i^T}}{v}    \sum_{k=1}^v{\widetilde{h^k(x_i)}}          \right)
\label{eq_11}
\end{equation}
where, 
\begin{equation}
W_i(m)=exp\left(    \frac{-{Y_i}^T}{v} \left( \sum_{k=1}^vf_k^{m-1}(x_i) \right)       \right)
\label{eq_12}
\end{equation}
${\widetilde{h^k(x)}} $ is local hypothesis vector on $m^{th}$ round on view space $k$ and $n$ is number of training examples.
$W_i(m)$ can be considered as the weight of $x_i$ on $m^{th}$ stage of boosting. Since $W_i(m)$ depends only on $f_k^{m-1}$, it is a constant for the optimization problem at the $m^{th}$ iteration. Following Eq.(\ref{eq_12}) we can write,
\begin{equation}
W_i(m+1)=exp\left(      \frac{-{Y_i}^T}{v} \left( \sum_{k=1}^vf_k^{m-1}(x_i)  +\beta\sum_{k=1}^v{\widetilde{h_k^m(x_i)}}               \right)           \right) 
\end{equation}
\begin{equation}
=W_i(m)exp\left(\frac{-\beta{Y_i^T} }{v} \sum_{k=1}^v{\widetilde{h_k^m(x_i)}}                                   \right) \label{eq_14}
\end{equation}
Eq.(\ref{eq_14}) is the weight update rule for our proposed SAMA-AdaBoost algorithm. Specifically, if an example $x_i$ has been misclassified on total $b_i$ views then following the steps of Eq.(\ref{eq_4}) it can be shown easily that the weight update rule is given by,
\begin{equation}
W_i(m+1)=W_i(m)exp\left[-\beta\left(1-\frac{2b_i}{v} \right)                              \right]
\end{equation}
We now return to our optimization objective as stated in Eq.(\ref{eq_11}). For simplicity we consider,
\begin{equation}
A=\sum_{i=1}^nW_i(m)exp\left(\frac{-\beta{Y_i^T}}{v}    \sum_{k=1}^v{\widetilde{h^k(x_i)}}          \right)
\label{eq_16}
\end{equation}
For illustration purpose, suppose that $x_1$ is misclassified on total $b_1$ views. Considering Eq.(\ref{eq_16}) only for $x_1$, we get

\begin{equation}
\begin{multlined}
A_1=W_1(m)exp\Big( \frac{\beta}{v}\sum_{b=1}^{b_1} \mathds{1}\left[ {Y_1^T\neq{\widetilde{h^b(x_1)}}}\right ] \\
-\frac{\beta}{v}\sum_{c=1}^{v-b_1} \mathds{1}\left[ {Y_1^T={\widetilde{h^c(x_1)}}}\right ]    \Big )
\end{multlined}
\end{equation} 

\begin{equation}
\begin{multlined}
=W_1(m)exp\Big( \frac{\beta}{v}\sum_{b=1}^{b_1} \mathds{1}\left[ {Y_1^T\neq{\widetilde{h^b(x_1)}}}\right ] \\
-\frac{\beta}{v}\sum_{c=1}^{v-b_1}(1- \mathds{1}\left[ {Y_1^T\neq{\widetilde{h^c(x_1)}}})\right ]    \Big )
\end{multlined}
\end{equation} 
In general if we consider this approach for all $x_i$ then we can rewrite Eq.(\ref{eq_16}) as follows,
\begin{equation}
\begin{multlined}
A=\sum_{i=1}^nW_i(m)exp\Big( \frac{\beta}{v}\sum_{b=1}^{b_i} \mathds{1}\left[ {Y_i^T\neq{\widetilde{h^b(x_i)}}}\right ]\\
+ \sum_{c=1}^{v-b_i} \mathds{1}\left[ {Y_i^T\neq{\widetilde{h^c(x_i)}}}\right ]-\frac{\beta}{v}(v-b_i) \Big)
\end{multlined}
\label{eq_19}
\end{equation} 
Note that $\mathds{1}\left[ {Y_i^T\neq{h_c(x_i)}}\right ]$ is identically zero because the index $c$ runs over weak learners which have correctly classified $x_i$. Thus, Eq.(\ref{eq_19}) reduces to,
\begin{equation}
\begin{multlined}
A=\sum_{i=1}^nW_i(m)exp\Big( \frac{\beta}{v}\sum_{b=1}^{b_i} \mathds{1}\left[ {Y_i^T\neq{\widetilde{h_b(x_i)}}}\right] -\frac{\beta}{v}(v-b_i) \Big)]
\end{multlined}
\label{eq_20}
\end{equation} 
To minimize $A$ in Eq.(\ref{eq_20}), we need a set of weak learners such that $\sum_{b=1}^{b_i} \mathds{1}\left[ {Y_i^T\neq{\widetilde{h_b(x_i)}}}\right]$ is minimal. Thus we have,
$\sum_{k=1}^v{\widetilde{h^{k*}(x_i)}}$=set of weak learners which manifest least possible exponential weighted error given by Eq.(\ref{eq_20}). With this optimal set of weak learners we now aim to evaluate the optimum value of $\beta$ i.e., $\beta^{m*}$. Rewriting Eq.(\ref{eq_20}) we get,
\begin{equation}
A=\sum_{i=1}^nW_i(m)exp\left[-\beta\left(1-\frac{2b_i}{v}    \right)      \right]
\label{eq_to_optimize}
\end{equation}
Differentiating $A$ w.r.t $\beta$ and setting to zero yields,
\begin{equation}
\sum_{i=1}^nW_i(m)exp\left( \frac{2\beta b_i}{v} \right)=\frac{2}{v}\sum_{i=1}^nW_i(m)exp\left( \frac{2\beta b_i}{v} \right)
\end{equation}
We solve numerically for $\beta^{m*}$ by optimizing $A$ of Eq.(\ref{eq_to_optimize}). The exact procedure to determine $\beta^{m*}$ is illustrated in the next section. Thus, $\sum_{k=1}^v{\widetilde{h^{k*}(x_i)}}$ and $\beta^{m*}$ represent the optimum set of weak learners and learning rate that needs to be updated in the additive model at $m^{th}$ iteration.
\subsection{Implementation of SAMA-AdaBoost}
In the previous subsection we presented the mathematical framework  of multiview assisted forward stagewise additive model of proposed SAMA-AdaBoost. Now we explain the steps for implementing SAMA-AdaBoost for any real life classification task.
\subsubsection{\textbf{Initial parameters}}
\begin{itemize}
\item Training examples $(x_1,y_1)$, $(x_2,y_2)$,....,$(x_n,y_n)$; \\$y_i \in \{1,2,,,K\}$
\item Total $V$ view/feature spaces 
\item Weak hypothesis $h_v^t(x)$ on $v^{th}$  view space on $t^{th}$ boosting round
\item $T$: total boosting rounds
\item Initial weight distribution $W^t(x_i)  $=  $\frac{1}{n}~~ \forall i \in\{1,2...n\}$
\end{itemize}
\subsubsection{\textbf{Communication across views and grading difficulty of training example}}
After a boosting round $t$, weak learners across views share their classification results. Let an example $x_i$ be misclassified over total $b_i^t$ views. Following the arguments in Eq.(\ref{eq_4}), difficulty of $x_i$ at boosting round $t$ is asserted by $\theta_t$,
\begin{equation}
\label{difficulty}
\theta_t=\frac{2b_i^t}{V}-1
\end{equation}

\subsubsection{\textbf{Weight update rule }}
\begin{itemize}
\item Learning rate $\beta_t$ is set to $\beta^{m*}$ which optimizes Eq.(\ref{eq_to_optimize}) after $t(=m)$ boosting rounds.
\item Weight update rule
 \begin{equation}
W^{t+1}(x_i)  =W^t(x_i).exp\left[-\beta_t\left(1-\frac{2b_i^t}{V}    \right)      \right]
\label{eq_weight_update}
\end{equation}
\end{itemize}
It is noteworthy that if $b_i^t=V$, then Eq.(\ref{eq_weight_update}) reduces to ,
\begin{equation}
W^{t+1}(x_i)  =W^t(x_i).exp(\beta_t)
\end{equation}
which is the usual weight update rule of traditional AdaBoost when $x_i$ has been misclassified. Similarly, when $b_i^t=0$,
\begin{equation}
W^{t+1}(x_i)  =W^t(x_i).exp(-\beta_t)
\end{equation}
which is the usual weight update rule of traditional AdaBoost when $x_i$ has been correctly classified. Thus our proposed algorithm is a generalization of AdaBoost and aids in asserting degree of difficulty of sample space instead of `1/-1' loss. The proposed weight update rule thus helps the learning algorithm to dynamically assert more importance to relatively "tougher" misclassified example compared to "easier" misclassified example.

\subsubsection{\textbf{Fitness measure of local weak learners}}
We first determine the fitness of a local weak learner, $h_v^t(x)$.
\begin{itemize}
\item Define a set $A_v^t$ such that,
\begin{equation}
A_v^t=\{ x_i| h_v^t(x_i) = y_i  \}
\end{equation}
\item Correct classification rate of $h_v^t(x)$ is given by,
\begin{equation}
r_v^t=\frac{|A_v^t|}{n}
\end{equation}
\end{itemize}
We argue that $r_v^t$ alone is not an appropriate fitness metric for $h_v^t(x)$. We found during experiments that there can be a weak learner $h_j^t(x)$ whose $r_j^t$ is low but it tends to correctly classify "tougher" examples. So, fitness of $h_v^t(x)$ should be evaluated not only based on $r_v^t$ but also based on difficulty of sample space which $h_v^t(x)$ correctly classifies.
\begin{itemize}
 \item Reward of $h_v^t(x)$ is determined by $R_v^t$ as follows,
\end{itemize}
$$R_v^t=\sum_{i: h_v^t(x_i)=y_i}W^t(x_i).|h_v^t(x_i)|$$
\begin{equation}
-\sum_{j:h_v^t(x_j)\neq y_j}\{1-W^t(x_j).|h_v^t(x_j)|\}
\end{equation}
\begin{itemize}
\item Finally, fitness of $h_v^t(x)$ is given by  $F_v^t$ as follows,
\end{itemize}
\begin{equation}
F_v^t=r_v^t(1+R_v^t)
\label{eq_reward}
\end{equation}
Eq.(\ref{eq_reward}) highly rewards the classifiers which correctly classify "tougher" examples with high confidence while highly penalizing weak learners which misclassify "easier" examples with high conviction.  Repeat steps 2-4 for $T$ times.
\subsubsection{\textbf{Conglomerating local weak learners}}
\begin{itemize}
\item SAMA-AdaBoost.V1:: In this version the final meta-classifier $H_f(x)$ is given by,
\end{itemize}
\begin{equation}
H_f(x)=\left \lfloor{\frac{\sum_{t=1}^T\sum_{v=1}^V F_v^t.h_v^t(x)}{V\sum_{t=1}^T\sum_{v=1}^VF_v^t} } \right \rfloor
\end{equation}
where $\lfloor{x}\rfloor$ represents nearest integer to (x).
\begin{itemize}
\item SAMA-AdaBoost.V2:: In this version the final meta-classifier $H_f(x)$ is given by,
\end{itemize}
\begin{equation}
H_f(x) = \underset{p\in \{1,2...,K\}} {\mathrm{argmax}}~\left[ \sum_{t=1}^T\sum_{v=1}^VF_v^t.|h_v^t(x)|_p \right]
\end{equation}

where, $|h_v^t(x)|_p$ is prediction confidence for class p.
\section{Study on Convergence Properties}
\subsection{\textbf{Error Bound on Training Set}}
In this section we derive an analytical expression which upper bounds training set error of multiview boosting and later we empirically compare the variations of the bounds of SAMA-AdaBoost, MA-AdaBoost at different levels of boosting. Without loss of generality, the analysis is performed on binary classification and we consider a simpler version of SAMA and MA-AdaBoost, which fuses weak multiview learners by simple majority voting instead of reward-penalty based voting. The motivation of the second simplification is to appreciate the difference of the core boosting mechanisms of the comparing three paradigms.
The final boosted classifier learned on $V$ views after $T$ boosting rounds is given by,
\begin{equation}
H_{fin}(x)=sign \left ( \sum_{v=1}^V\sum_{t=1}^T \beta_t h^t_v(x) \right )
\label{conv_hfinal}
\end{equation}
We define $F(x)$ as,
\begin{equation}
F(x)= \left ( \sum_{v=1}^V\sum_{t=1}^T \beta_t h^t_v(x) \right )
\end{equation}
A normalized version of weight update rule for SAMA-AdaBoost can be written as,
\begin{equation}
W^{t+1}(x_i)=\frac{W^t(x_i)\exp \left \{-\beta_t \left (1-\frac{2b^t_i}{V}\right ) \right \}}{Z^t}
\label{eq_conv_weight_update}
\end{equation}
where normalization factor, $Z^t$ is given by,
\begin{equation}
Z^t=\sum_{i=1}^m W^t(x_i)\exp \left \{-\beta_t \left (1-\frac{2b^t_i}{V}\right ) \right \}
\end{equation}
The recursive nature of Eq.(\ref{eq_conv_weight_update}) enables us write the final weight on $x_i$, $W^{fin}(x_i)$ as,
\begin{equation}
W^{fin}(x_i)=\frac{ \exp \left \{\sum_{t=1}^T-\beta_t \left (1-\frac{2b^t_i}{V}\right ) \right \}}{m\prod_{t=1}^TZ^t}
\end{equation}
\begin{equation}
=\frac{\exp \left [ \sum_{t=1}^T\beta_t  \{ 1-\frac{2(\sum_{v=1}^V(1-y_ih^t_v(x_i)))}{V}       \} \right ] }{m\prod_{t=1}^TZ^t}
\end{equation}
\begin{equation}
\implies W^{fin}(x_i)=\frac{\exp  \left (\sum_{t=1}^T\beta_t -\frac{2y_i F(x_i)}{V} \right )}{m\prod_{t=1}^TZ^t}
\end{equation}
\begin{equation}
\implies \exp  \{ -y_i F(x_i) \}^\frac{2}{V} = \frac{m W^{fin}\prod_{t=1}^TZ^T}{\exp (\sum_{t=1}^T\beta_t)}
\label{eq_implication}
\end{equation}
Now, training set error incurred by $H_{fin}(x)$ can be represented as,
\begin{equation}
\frac{1}{m}\sum_{i}  \begin{cases}
1 & if ~~y_i \neq H_{fin}(x_i)\\
0 & if ~~y_i = H_{fin}(x_i)\\
\end{cases}
\end{equation}
\begin{equation}
\leq \frac{1}{m}\sum_{i=1}^m\exp (-y_i F(x_i))^\frac{2}{V}
\end{equation}
\begin{equation}
=\frac{1}{m}\frac{m W^{fin}\prod_{t=1}^TZ^t}{\exp (\sum_{t=1}^T\beta_t)}
\end{equation}
\begin{equation}
=\frac{\prod_{t=1}^TZ^t}{\exp(\sum_{t=1}^T\beta_t)}
\label{eq_train_bound}
\end{equation}

\par Eq.(\ref{eq_train_bound}) provides an upper bound for multiview boosting paradigms such as SAMA and MA-AdaBoost. It is to be remembered that though Eq.(\ref{eq_train_bound}) holds true for both SAMA-AdaBoost and MA-AdaBoost, $\beta_t$, and thus, explicitly $Z^t$, are different for the two algorithms. In Table \ref{table_train_bound} we report the upper bounds calculated for SAMA-AdaBoost, MA-AdaBoost at different levels of boosting on eye classification task (Refer to Section \ref{section_eye} for dataset and implementation details). A lower error bound is an indication that the ensemble has learnt the examples on the training set and is less susceptible to train set misclassification. The exact values in Table \ref{table_train_bound} are not important but the scales of the magnitudes are worth noticing. We see that ensemble space of proposed SAMA-AdaBoost is able to learn much faster compared to MA-AdaBoost. The rate of decrease of error bound is aggressively faster with each round of boosting for proposed SAMA-AdaBoost compared to MA-AdaBoost. We see that error bound for SAMA-AdaBoost suffers a lofty drop from the order of $10^{-7}$ to $10^{-32}$ when training is increased from 15 to 20 rounds of boosting. On contrast, error bound of MA-AdaBoost reduces insignificantly and stays at $10^{-5}$. Table \ref{table_train_bound} is a strong indication systematic optimization of SAMA-AdaBoost's loss function fosters in faster convergence rate on training set.

\begin{table}
	\caption{Comparison of training set error bounds (Eq.(\ref{eq_train_bound})) after different levels of boosting (T). A lower value of error bound signifies that an ensemble is prone is make less error on the training set. }
	\centering
	\begin{tabular}{c c c }
		\hline\hline\\
		Boosting Rounds: T& \textbf{SAMA-AdaBoost:Proposed}&MA-AdaBoost\\
		5&7.1*$10^{-1}$&7.6*$10^{-1}$\\
		10&3.2*$10^{-4}$&0.9*$10^{-2}$\\
		15&8.3*$10^{-7}$&4.2*$10^{-4}$\\
		20&5.0*$10^{-32}$&1.8*$10^{-5}$\\
		25&2.5*$10^{-33}$&1.1*$10^{-5}$\\

		\hline
		\label{table_leaf_test}
		\label{table_train_bound}
	\end{tabular}
\end{table}
\begin{figure*}[!t]
	\centering
	\subfigure {\includegraphics[width=0.33\linewidth]{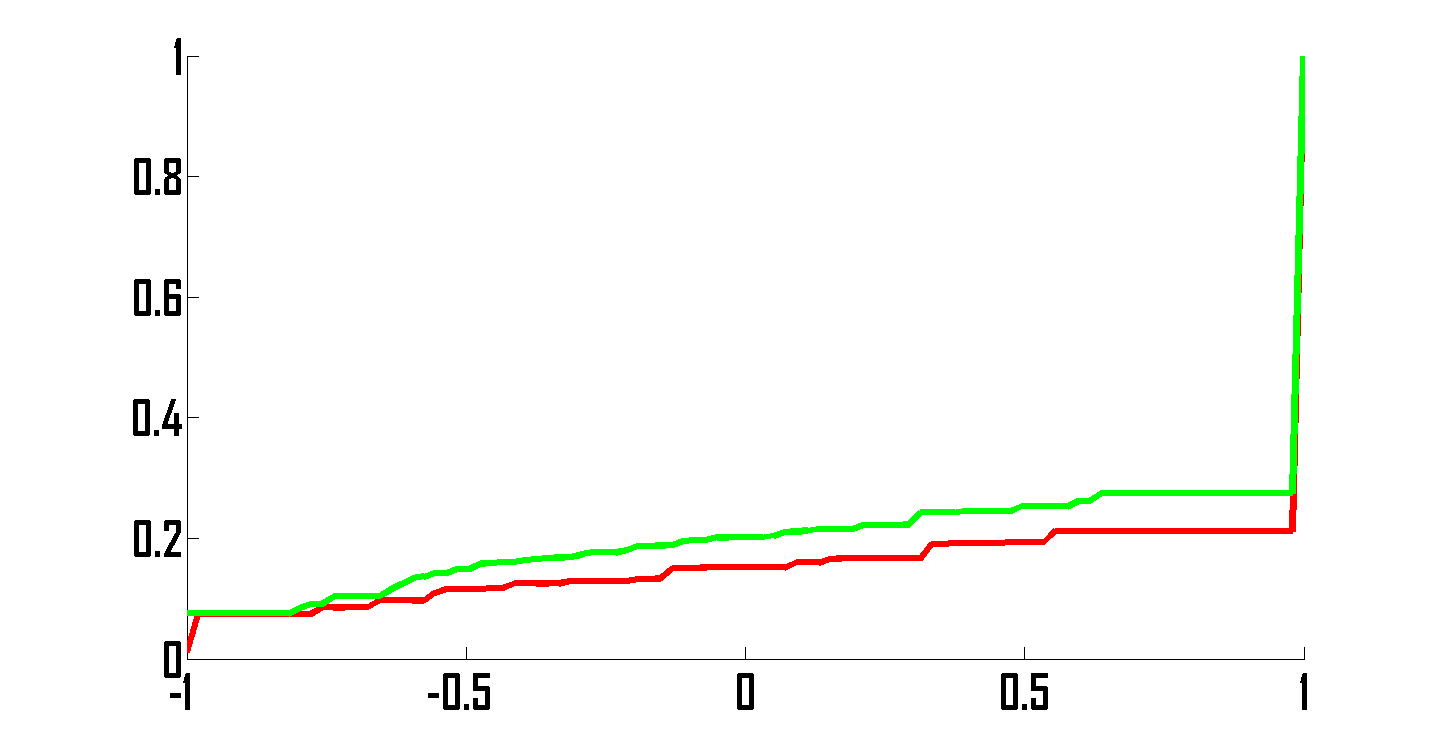}}\hfill \hspace*{-0.9em}
	\subfigure {\includegraphics[width=0.33\linewidth]{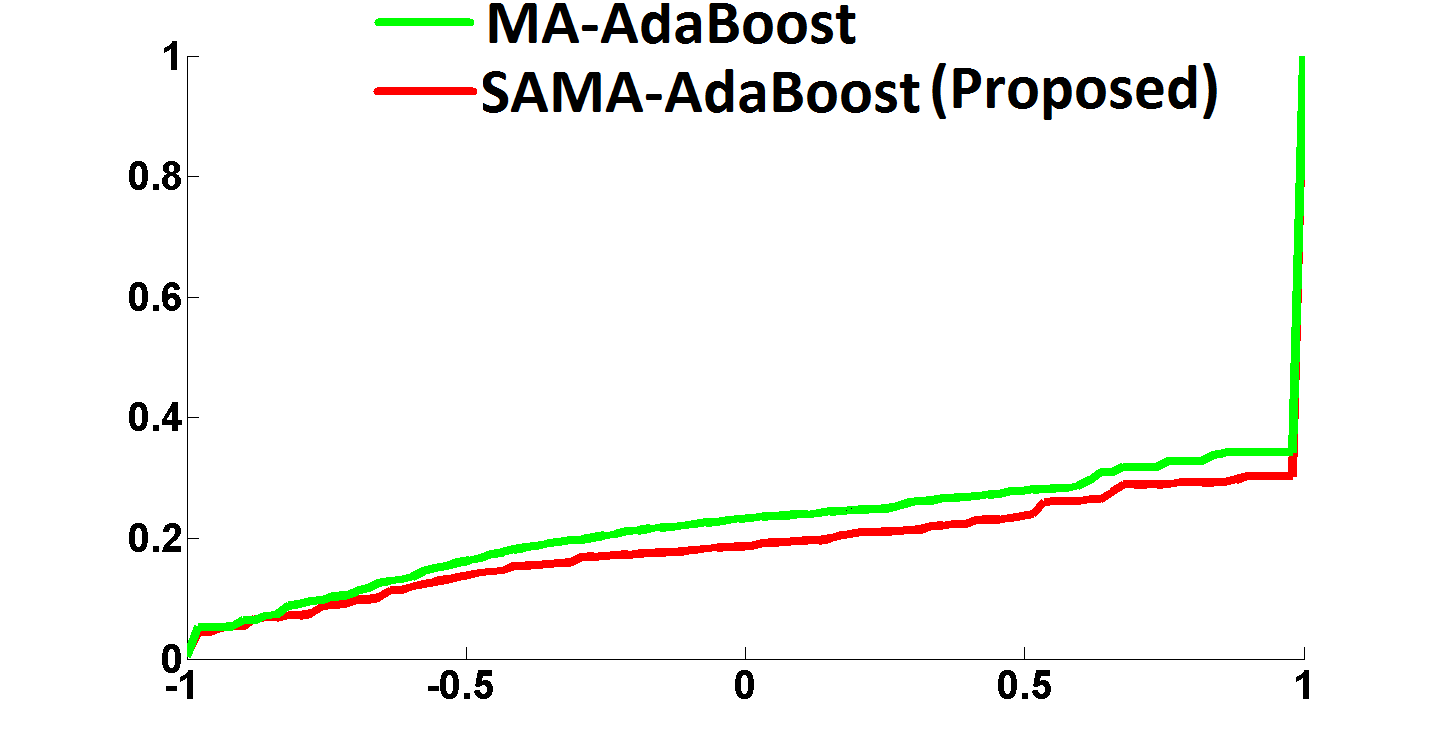}}\hfill \hspace*{-0.9em}
	\subfigure {\includegraphics[width=0.33\linewidth]{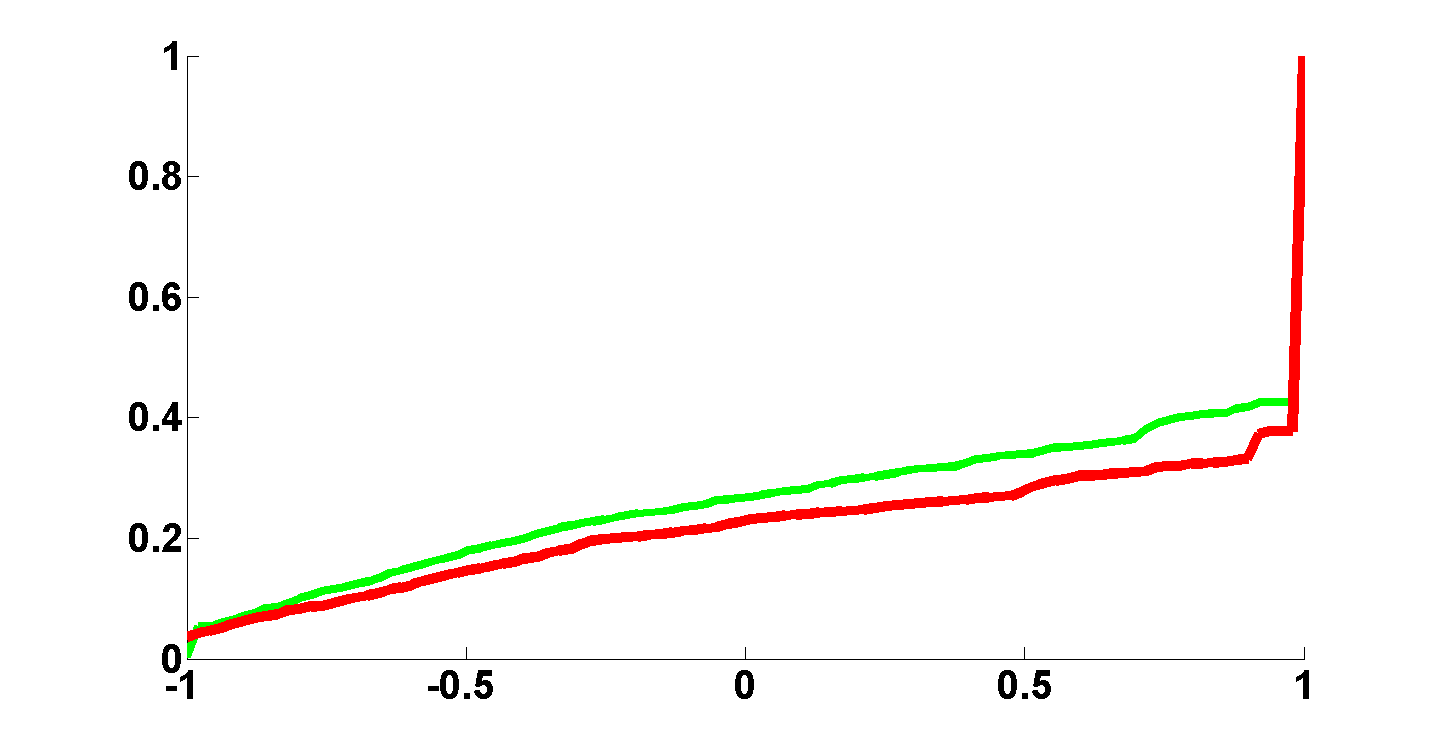}}\hfill \\
	
	\caption{Margin distribution graphs on 100-Leaves classification task after 5,10 and 15 rounds of boosting. Vertical axis denotes the fraction of training sample space having margin $\leq x$ (horizontal axis).}
	\label{fig_margin_distribution}
\end{figure*}
\subsection{\textbf{Generalization Error and Margin Distribution Analysis}}
\subsubsection{Visualizing Margin Distribution}For understanding generalization property of boosting, training set performance reveals only a part of the entire explanation. It has been shown in \cite{margin_boost2} that more confidence on training set explicitly improves generalization performance. Frequently, margin on training set is taken as the metric of confidence of boosted ensemble. In the context of boosting, margin is defined as follows: Suppose that the final boosted classifier is a convex combination of base/weak learners. Weightage on a particular class for a training example, $x_i$ is taken as summation of the convex weights of the base learners. Margin for $x_i$ is computed as difference of weight assigned to the correct label of $x_i$ and the highest weight assigned to an incorrect label. Thus, margin spans over the range $\in [-1,1]$. It is easy to see that for a correctly classified $x_i$, margin is positive while it is negative in case of misclassification. Significantly high positive margin manifests greater confidence of prediction. It has been shown in \cite{margin_boost2,margin_boost1} that for high generalization accuracy it is mandatory to have minimal fraction of training example with small margin. Margin distribution graphs are usually studied in this regard. A margin distribution graph is a plot of the fraction of training examples with margin atmost $\psi$ as a function of $\psi \in [-1,1]$. In Fig. \ref{fig_margin_distribution} we analyze the margin distribution graphs of SAMA-AdaBoost and MA-AdaBoost. We have used the same simulation setup on the 100-Leaves classification task as will be discussed in Section \ref{subsection_100_leaves}.
\par Consistently, we find that the margin distribution graph of SAMA-AdaBoost lies below that of MA-AdaBoost. Such a distribution means that given a margin, $\psi_m$, SAMA-AdaBoost always tends to have fewer examples with margin $\leq \psi_m$ compared to MA-AdaBoost. This explicitly makes ensemble space of SAMA-AdaBoost more confident on training set and thereby manifesting superior performance on test set.\\

\subsubsection{Bound on Margin Distribution}
In this section we provide an analytical expression (on a similar note to \cite{margin_boost2}) for estimating the upper bound of margin distribution of an ensemble space created by SAMA-AdaBoost and MA-AdaBoost. Later, we show through numerical simulations that boosting inherently encourages to decrease fraction of training example with low margin as we keep on increasing the number of boosting rounds. Let, $\mathbf{X}$, $\mathbf{Y}$ denote instance and label space respectively and training examples are generated according to some unknown but fixed distribution over $\mathbf{X}\times \{-1,1\}$. $\Delta$ denote the training set consisting of $m$ ordered pairs, i.e., $\Delta=\{(x_1,y_1), (x_2,y_2),....(x_m,y_m)\}$, chosen according to that same distribution. Define, $\mathbf{P}_{x,y \thicksim \Delta} [\Phi]$ as the probability of event $\Phi$ given that the example $(x,y)$ has been randomly drawn from $\Delta$ following a normal distribution. Under unambiguous context, $\mathbf{P}_{x,y \thicksim \Delta} [\Phi]$ and $\mathbf{P}_{\Delta} [\Phi]$ are used interchangeably. Similarly, $\mathbf{E}_\Delta [\Phi]$ refers to the expected value. $\mathscr{H}$ is defined as the convex combination of the boosted base learners.
\begin{equation}
\mathscr{H}(x)=\frac{\sum_{t=1}^T \sum_{v=1}^V \beta_t h^t_v(x)}{\sum_{t=1}^T \beta_t}
\label{eq_convex_final}
\end{equation}
Given, $\theta: 0 \leq \theta \leq 1$, we are interested to find an upper bound on,
\begin{equation}
\mathbf{P}_{(x,y) \thicksim \Delta}[y\mathscr {H}(x) \leq \theta]
\end{equation}
If we assume, $y\mathscr {H}(x) \leq \theta$, it implies,
\begin{equation}
y \sum_{t=1}^T \sum_{v=1}^V \beta_t h^t_v(x) \leq \theta \sum_{t=1}^T \beta_t
\end{equation}
\begin{equation}
\implies \exp \left (-y \sum_{t=1}^T \beta_t \sum_{v=1}^V h^t_v(x) + \theta\sum_{t=1}^T \beta_t   \right) ^\frac{2}{V \sum_{t=1}^T \beta_t }  \geq 1
\end{equation}
$\implies \mathbf{P}_{(x,y) \thicksim \Delta} [y\mathscr {H}(x) \leq \theta] $
\hspace{-2mm}\begin{equation}
\leq  \mathbf{E}_{(x,y) \thicksim \Delta} \left [ \exp \left (-y \sum_{t=1}^T \beta_t \sum_{v=1}^V h^t_v(x) + \theta\sum_{t=1}^T \beta_t   \right) ^\frac{2}{V \sum_{t} \beta_t } \right ]
\end{equation}
\begin{equation}
= \frac{\exp (\theta \sum_{t}\beta_t)^{\frac{2}{V\sum_{t}\beta_t}}}{m} \sum_{i=1}^m \exp   (-y_i \sum_{t}\beta_t \sum_{v}h^t_v(x_i)    )^{\frac{2}{V\sum_{t}\beta_t}}
\end{equation}
\begin{equation}
=\frac{\exp (\theta)^\frac{2}{V}}{m} \sum_{i=1}^m \exp \left ( -y_i \mathscr{H}(x_i)  \right )^\frac{2}{V}
\end{equation}
On a similar argument presented in Eq.(\ref{eq_implication}), we can write,
\begin{equation}
\mathbf{P}_{(x,y) \thicksim \Delta} [y\mathscr {H}(x) \leq \theta] \leq \frac{\exp (\theta)^\frac{2}{V}}{m} \frac{\prod_{t=1}^TZ^t}{\exp (\sum_{t=1}^T\beta_t)}
\label{eq_margin_bound}
\end{equation}
\begin{figure}
	\centering
	\includegraphics[scale=0.25]{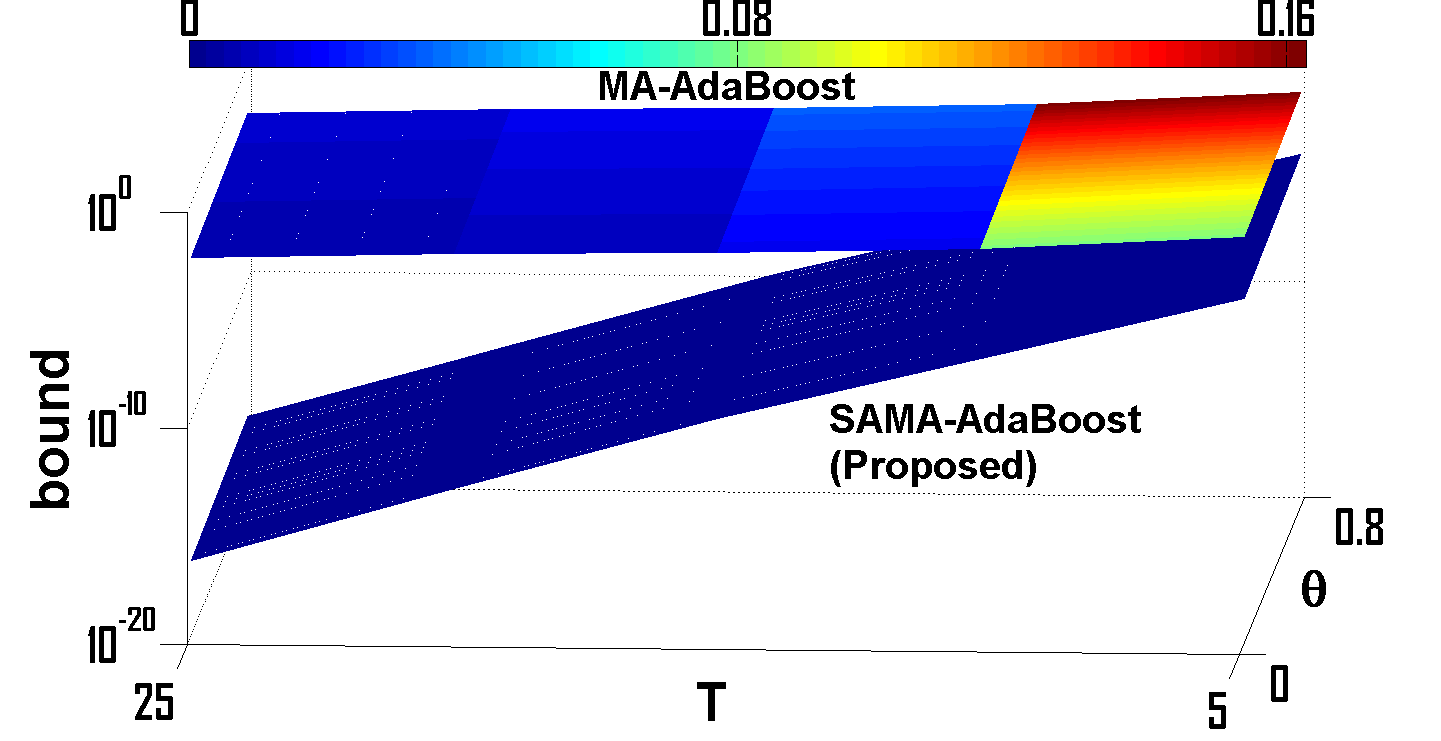}
	\caption{Variation of upper bound of margin distribution on eye classification dataset after different rounds $(T)$ of boosting. Bound represents the upper bound of probability of sampling a training example with margin $\leq \theta$. }
	\label{fig_margin_bound_analysis}
\end{figure}
Eq.(\ref{eq_margin_bound}) gives an upper bound of sampling training examples with margin $\leq \theta$. Intuitively, we want this probability to be less because that aids in margin maximization. In Fig. \ref{fig_margin_bound_analysis} we illustrate the variation of this bound at different levels $(T)$ of boosting for SAMA-Boost and MA-AdaBoost on eye classification dataset (refer Section \ref{section_eye}). In the figure, bound represents the upper bound of probability of sampling a training example with margin $\leq \theta$, i.e., $\mathbf{P}_{(x,y) \thicksim \Delta} [y\mathscr {H}(x) \leq \theta]$. For both SAMA-Boost and MA-Boost, at a given $T$, we observe that the bound decreases with decrease in $\theta$. This indicates that the ensembles discourage to possess training examples with low margin. Also, for a given $\theta$, the bound decreases with increase of $T$; the observation indicates that increasing rounds of boosting implicitly reduces existence of low margin examples.  A significant observation is that the decay rate of upper bound with $T$ for  SAMA-Boost is appreciably higher compared to that of MA-AdaBoost. Specifically, after 5 rounds of boosting,  upper bound for SAMA-Boost is $2.1\times10^{-4}$, while for MA-AdaBoost, the bound is $1.6\times10^-1$.  After 25 rounds of boosting, upper bound for SAMA-Boost is $5.4\times10^{-17}$, while for MA-AdaBoost, the bound is $6.3\times10^-2$. Analysis of this section thereby bolsters our claim that the learning rate of  proposed SAMA-AdaBoost algorithm is much faster compared to MA-AdaBoost's rate. Emsemble space of SAMA-AdaBoost manifests significantly lower probability of possessing low margin examples compared to that of MA-AdaBoost. Such observation guarantees better generalization capability for SAMA-AdaBoost. Empirical results in Section \ref{section_eye} will further strengthen our claim.
\section{Experimental Analysis}
In this section we compare our proposed SAMA-AdaBoost on challenging real world datasets with recent state-of-art collaborative and variants of non-collaborative traditional boosting algorithms. It has been shown in \cite{accv} that ".V2" version of MA-AdaBoost performs slightly better than ".V1" version and thus here we present results using SAMA-AdaBoost.V2 and MA-AdaBoost.V2.
\begin{figure}
\centering
\includegraphics[scale=0.22]{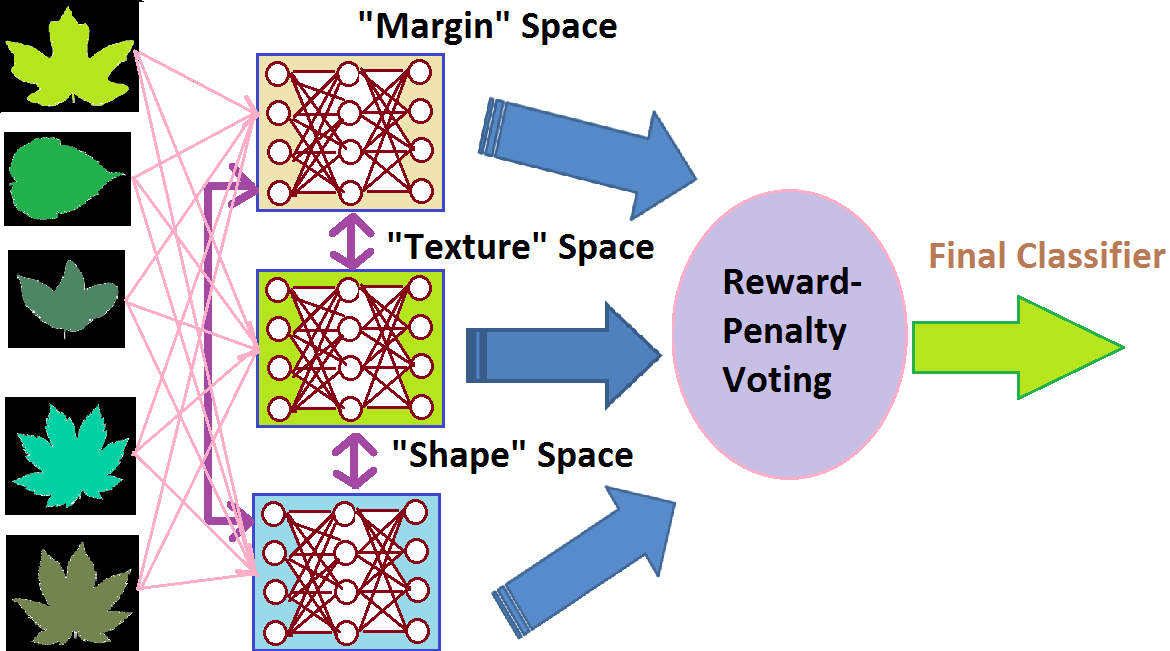}
\caption{A pictorial representation of our proposed multiview learning. On the extreme left we show extracted leaf segments of five out of one hundred classes of leaves from "100 leaves database \cite{leaves}".  Each extracted leaf segment is represented and learnt over three feature spaces with 2-layer ANN. Small bubbles in each rectangular box are drawn to mimic a 2-layer ANN architecture. Bidirectional pink arrows indicate communication across views and thereby performing collaborative learning. Finally, weak learners over different view spaces are combined by reward-penalty based voting. }
\label{fig_architecture}
\end{figure}
\vspace{-3mm}\subsection{100 Leaves Dataset \cite{leaves}}
\label{subsection_100_leaves}
This is a challenging dataset where the task is to classify 100 classes of leaves based on shape, margin and texture features. Each feature space is 16-dimensional with 16 examples per class. Such a heterogeneous feature set is apt to be applied on any multiview learning algorithm. For simulation purpose we have taken 2-layer ANN with 5 units in the hidden layer as baseline learner in each boosting round over each view space. The dataset is randomly shuffled and then split into 60:20:20 for training, validation and testing respectively. Regularization parameter $\lambda$ is selected by 5-fold validation. In Fig.\ref{fig_architecture} we pictorially represent the setting for our proposed multiview learning framework.


\begin{figure*}
\centering
\subfigure {\includegraphics[scale=.2]{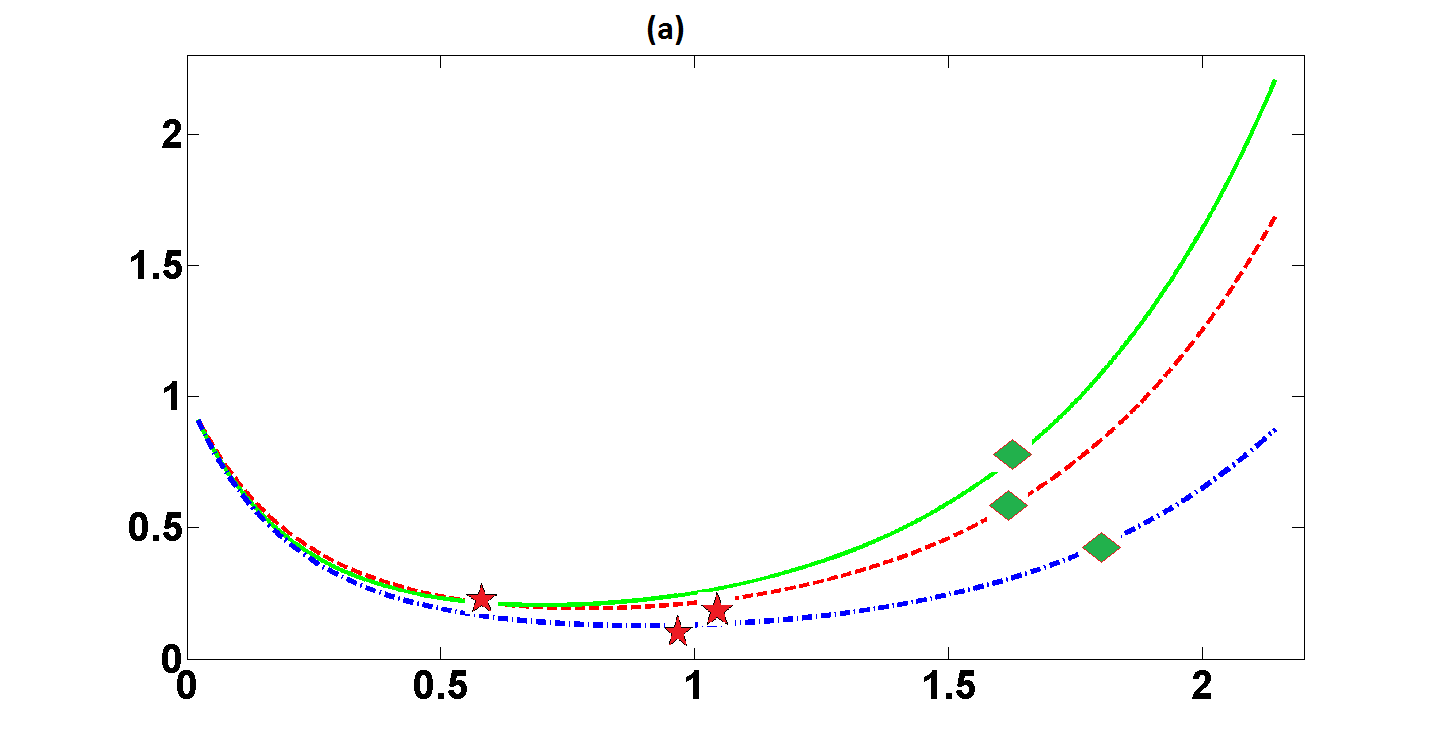}}
\subfigure {\includegraphics[scale=.2]{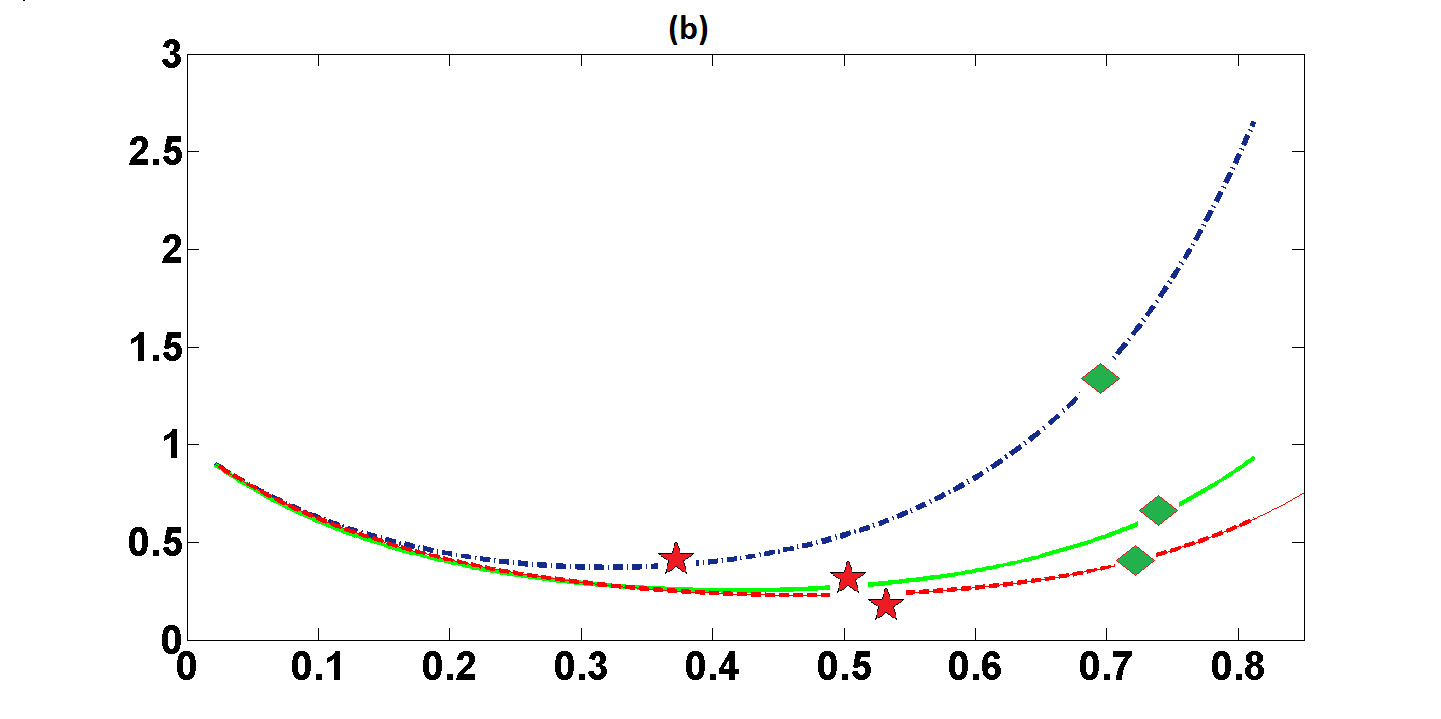}}
\subfigure {\includegraphics[scale=.2]{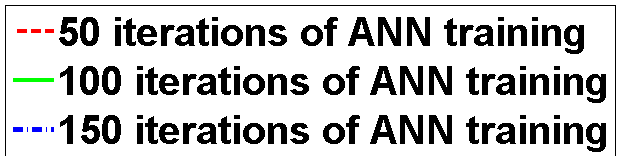}}
\subfigure {\includegraphics[scale=.2]{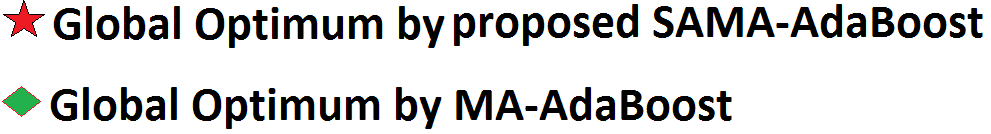}}\\
\subfigure {\includegraphics[scale=.2]{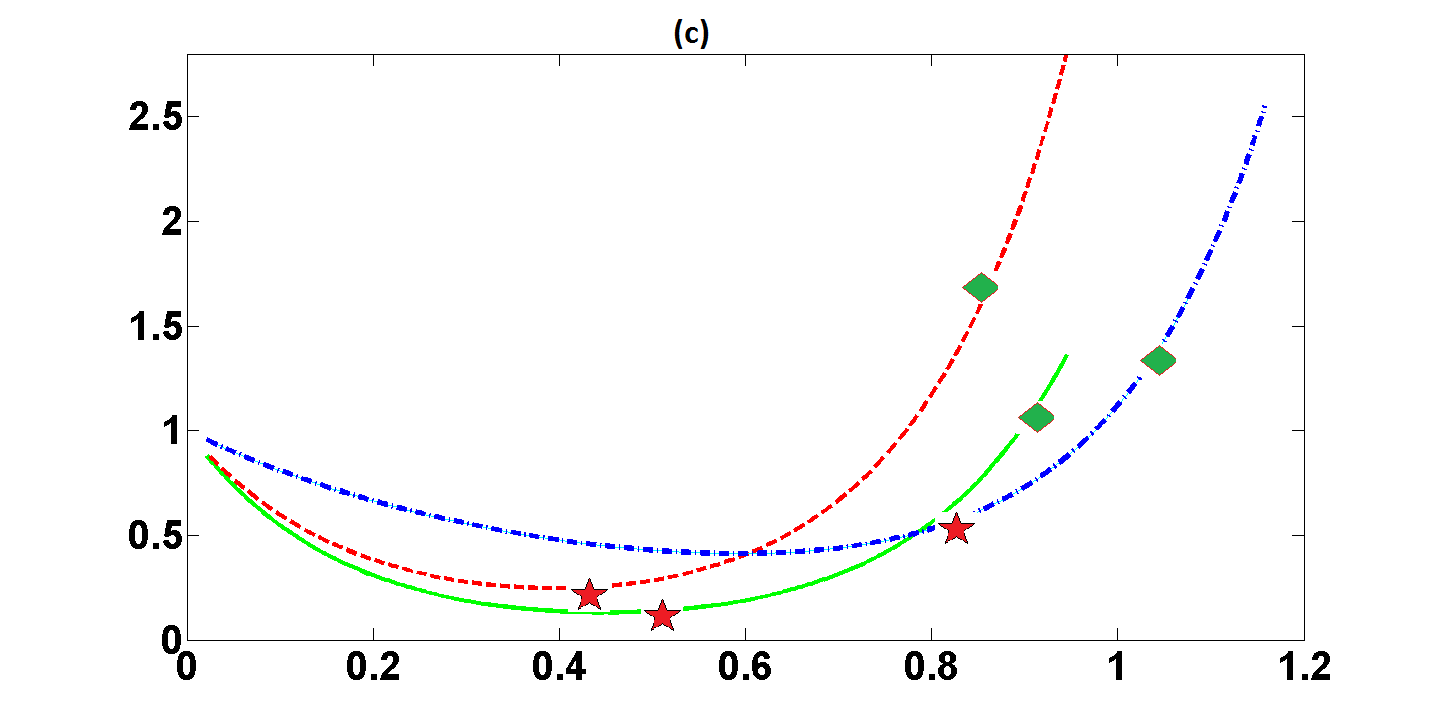}}
\subfigure {\includegraphics[scale=.2]{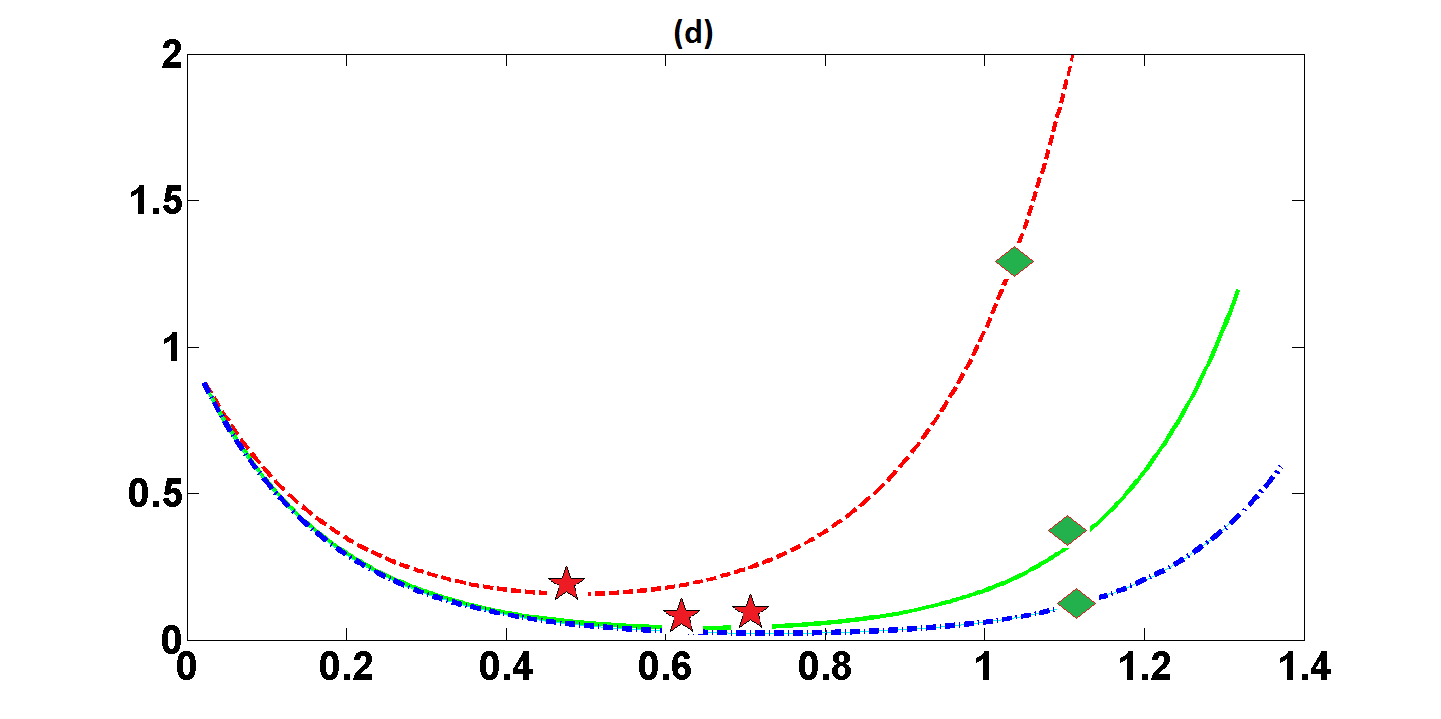}}\\
\caption{Plot of objective function $A$ (y-axis) versus $\beta$ (x-axis) as mentioned in Eq.(\ref{eq_to_optimize}) for "100 leaves dataset \cite{leaves}". Fig. (a) , (b), (c) and (d) represents  network trained over 2, 5, 7 and 10 rounds of boosting respectively. In each case we vary the number of ANN training iterations per boosting round as indicated by the colored lines. We see that our previous algorithm i.e., MA-AdaBoost fails to attain the global minimum whereas the proposed framework is able to localize very close to global minimum.}
\label{fig_beta}
\end{figure*}

\subsubsection{\textbf{Determining optimum value of }${\beta_t}$ }

In this section we illustrate the procedure to compute the optimum $\beta_t$ for minimizing $A$ in Eq.(\ref{eq_to_optimize}). It has been shown by Schapire \cite{explain_adaboost}, that the exponential loss incorporated in AdaBoost is strictly convex in nature and is void of local minima. In Fig.\ref{fig_beta} we plot $A$ versus $\beta_t$. As it can see seen that the functional variation of $A$ w.r.t $\beta_t$ is indeed convex in nature and thus we apply gradient descent and select that value of $\beta_t$ for which the gradient of the function is close to zero. Absence of local minima guarantees that we will converge near to the global (single) minima. In \cite{accv}, we naively evaluated $\beta_t$ as,
\begin{equation}
\beta_t=0.5\times \log \left [ \frac{1-\prod_{v=1}^VP_{W^t}(\mathbf{h_v^t(x_i)}\neq \mathbf{Y_i})}{\prod_{v=1}^VP_{W^t}(\mathbf{h_v^t(x_i)}\neq \mathbf{Y_i})} \right ]
\end{equation}
We mark the optimum locations evaluated by our algorithm with red stars in Fig. \ref{fig_beta}. We also mark the corresponding optimal points (green rhombus) evaluated using our previously proposed MA-AdaBoost\cite{accv} and it is evident that MA-AdaBoost fails to attain the global minimum. In Table \ref{table_beta_compare} we report the ratio of global optimum indicated by SAMA-AdaBoost and MA-AdaBoost to the actual global minimum of $A-\beta$ space. After two, five and ten  rounds of boosting, average ratios for SAMA-AdaBoost are 1.07, 1.02 and 1.03 respectively while the average ratios for MA-AdaBoost are 2.9, 1.6 and 3.5 respectively. We thus argue that MA-AdaBoost fails to localize at global minimum by significant margin compared to SAMA-AdaBoost and as a consequence, SAMA-AdaBoost has faster training set error convergence rate compared to MA-AdaBoost as we shall see shortly. Similar nature of $A-\beta$ dependency is observed on other datasets.
\begin{table}
	\caption{Ratio of global minimum evaluated by competing algorithms to the actual global minimum of $A-\beta$ space on "100 leaves dataset". The closer the ratio is to unity the better. \textbf{T}: total boosting rounds. \textbf{Iterations}: number of times ANNs are trained per boosting round. We note that the proposed SAMA-AdaBoost converges much closer to actual minimum compared to our previous work of MA-AdaBoost.}
	\centering
	\begin{tabular}{c c c c}
		\hline\hline\\
		T&Iterations&\textbf{SAMA-AdaBoost (Proposed)}&MA-AdaBoost\cite{accv}\\
		\hline\\
		&50&1.05&4.00\\
		2&100&1.14&2.32\\
		&150&1.02&1.30\\\\
		&50&1.06&1.21\\
		5&100&1.01&1.45\\
		&150&1.02&2.81\\\\
		&50&1.01&4.12\\
		10&100&1.05&2.52\\
		&150&1.08&1.26\\
		\hline
		\label{table_beta_compare}
	\end{tabular}
\end{table}
\subsubsection{\textbf{Comparison of classification performances}}
In this section we report the training and generalization performances of several boosted classifiers. For comparing with other boosting algorithms with ANN as baseline, we used the boosting framework as proposed in \cite{neural_boost}. For comparing with \cite{prob_adaboost} we have taken the sample fraction $f$=0.5 and the correction factor equals to 4 as indicated by the authors. We cannot compare our results with \cite{coadaboost,boost2} because these algorithms only support 2-class problems.

\begin{figure}[!b]
\includegraphics[scale=0.2]{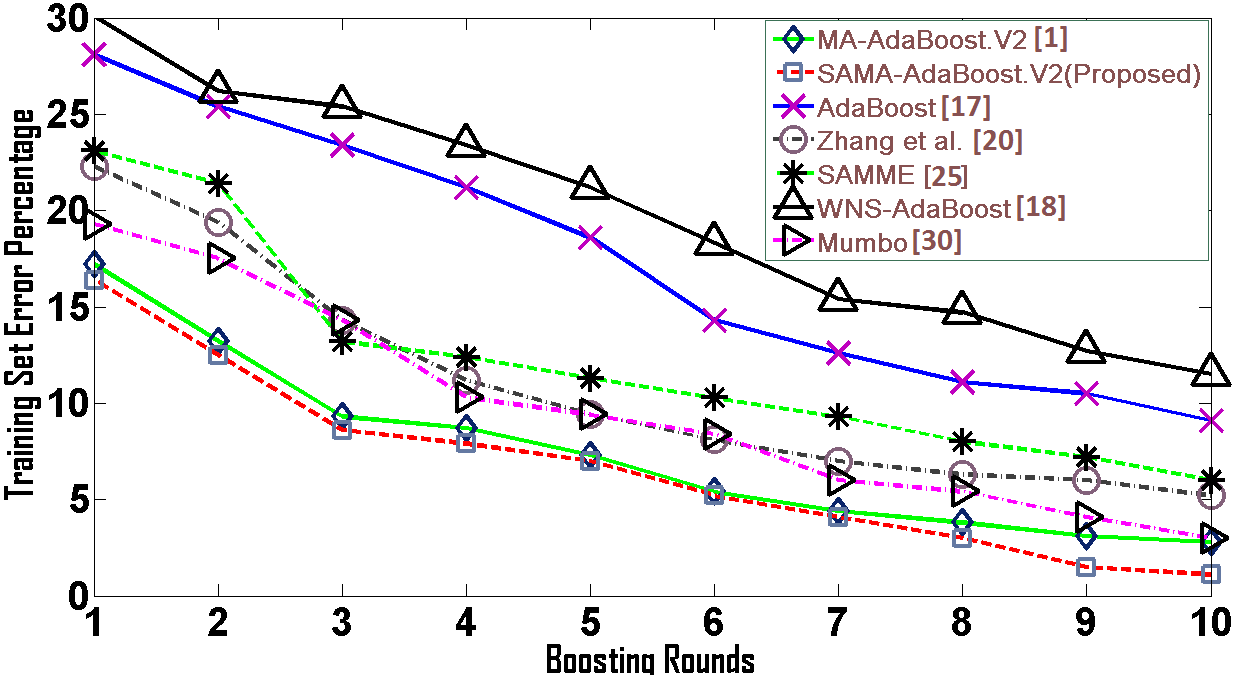}
\caption{Comparison of rates of convergence of training set error of different boosted classifiers on 100 Leaves dataset. From the graph it is evident that our proposed SAMA-AdaBoost has the fastest convergence rate. We start by training baseline ANNs with 50 iterations/round and increment upto 200 iterations/round in step of 20 and we measure misclassification rates in each step. In this figure we report the average results. }
\label{fig_leaf_train}
\end{figure}

\begin{table*}
\caption{Test Set Accuracy Percentages on "100 leaves dataset" of Competing Boosted Classifiers.\textbf{T:} total boosting rounds. \textbf{Iterations:} Number of back-propagation passes for training a ANN network per boosting round. }
\centering
\begin{tabular}{c c c c c c c c c}
\hline\hline\\
T&Iterations&\textbf{SAMA-AdaBoost:(Proposed)}&MA-AdaBoost \cite{accv}&Mumbo \cite{mumbo_speech} & SAMME \cite{samme} &AdaBoost \cite{adaboost_mh}  & Zhang \cite{prob_adaboost} & WNS \cite{wns}\\
\hline\\
&50&74.3&71.2&71.0&75.1&68.1&76.1&67.1\\
2&100&75.3&73.4&72.4&76.8&70.2&77.2&69.2\\
&150&76.9&75.4&73.9&77.4&71.2&78.2&70.4\\\\

&50& 86.2&83.1&82.8&80.4&77.4&79.8&76.1\\
5&100&90.3&87.2&85.3&82.3&79.8&82.0&78.4\\
&150&93.2&91.0&88.2&84.8&81.2&83.9&79.8\\\\

&50&97.2&95.9&93.2&89.1&87.4&90.8&86.4\\
10&100&98.4&96.2&94.8&90.2&89.8&92.3&87.2\\
&150&99.6&98.1&96.2&93.4&91.3&94.6&89.4\\
\hline
\label{table_leaf_test}
\end{tabular}
\end{table*}

\par In Fig. \ref{fig_leaf_train} we compare rate of convergence on training set error by the competing algorithms. Fig. \ref{fig_leaf_train} bolsters the boosting nature of our proposed algorithm because the training set error rate decreases with increase in number of boosting rounds. 
It is interesting to note that collaborative algorithms such as SAMA-AdaBoost, Mumbo and MA-AdaBoost perform worse compared to SAMME at low boosting rounds. Weak learner on each view space in collaborative algorithms is provided with only a subset of entire feature space. So  at low boosting rounds, weak learners are poorly trained and overall group performance is worsened. Conversely, SAMME is trained on entire concatenated feature space and even with low boosting rounds, weak learners of SAMME are superior compared to weak learners of collaborative algorithms.With increase of boosting rounds, performances of collaborating algorithms are enhanced compared to non-collaborative boosting frameworks. It is to be noted that the rate of convergence of training set error of SAMA-AdaBoost is faster compared to MA-AdaBoost and this is attributed to proper localization of  minimum in $A-\beta$ space by SAMA-AdaBoost. On average, SAMA-AdaBoost outperforms MA-AdaBoost, SAMME, Mumbo, AdaBoost, Zhang et al. and WNS-AdaBoost by margins of  3.8\%, 7.8\%, 4.3\%,10.2\%, 9.8\% and 11.2\% respectively.
\par Next, in Table \ref{table_leaf_test} we report the generalization error rates of the competing boosted classifiers. Our proposed algorithm achieves a classification accuracy rate of 99.6\% after 10 rounds of boosting with 150 iterations of ANN training per boosting round. The previously reported best result was 99.3\% by \cite{leaves} using probabilistic k-NN. On average, proposed SAMA-AdaBoost outperforms MA-AdaBoost, Mumbo, SAMME, AdaBoost, Zhang et al. and WNS-AdaBoost by margins of 2.3\%, 4,2\%, 5.3\%, 8.7\%, 4.8\% and 9.4\% respectively.

\subsection{Discriminating Between Eye and Non Eye Samples}
\label{section_eye}
\begin{figure}
\centering
\includegraphics[scale=0.5]{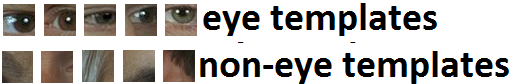}
\caption{Eye and non eye templates extracted from human face for 2-class classification problem. }
\label{fig_eye_example}
\end{figure}

In this section we compare our algorithm on a 2-class visual recognition problem. The task is discriminate human eye samples from non eye samples \cite{icvgip}. For simulation purpose we manually extracted 32$\times$32 eye and non eye templates from randomly chosen human faces from the web. Few examples are shown in Fig.\ref{fig_eye_example}. A training example is represented over two view spaces, viz. We utilized the two view representation as illustrated in \cite{icvgip}
The feature spaces are:
\begin{itemize}
	\item \textbf{Features from SVD-HSV space: 96D}
	\item \textbf{Features from SVD-Haar space: 48D}
\end{itemize}

\begin{figure}
\centering
\includegraphics[scale=0.65]{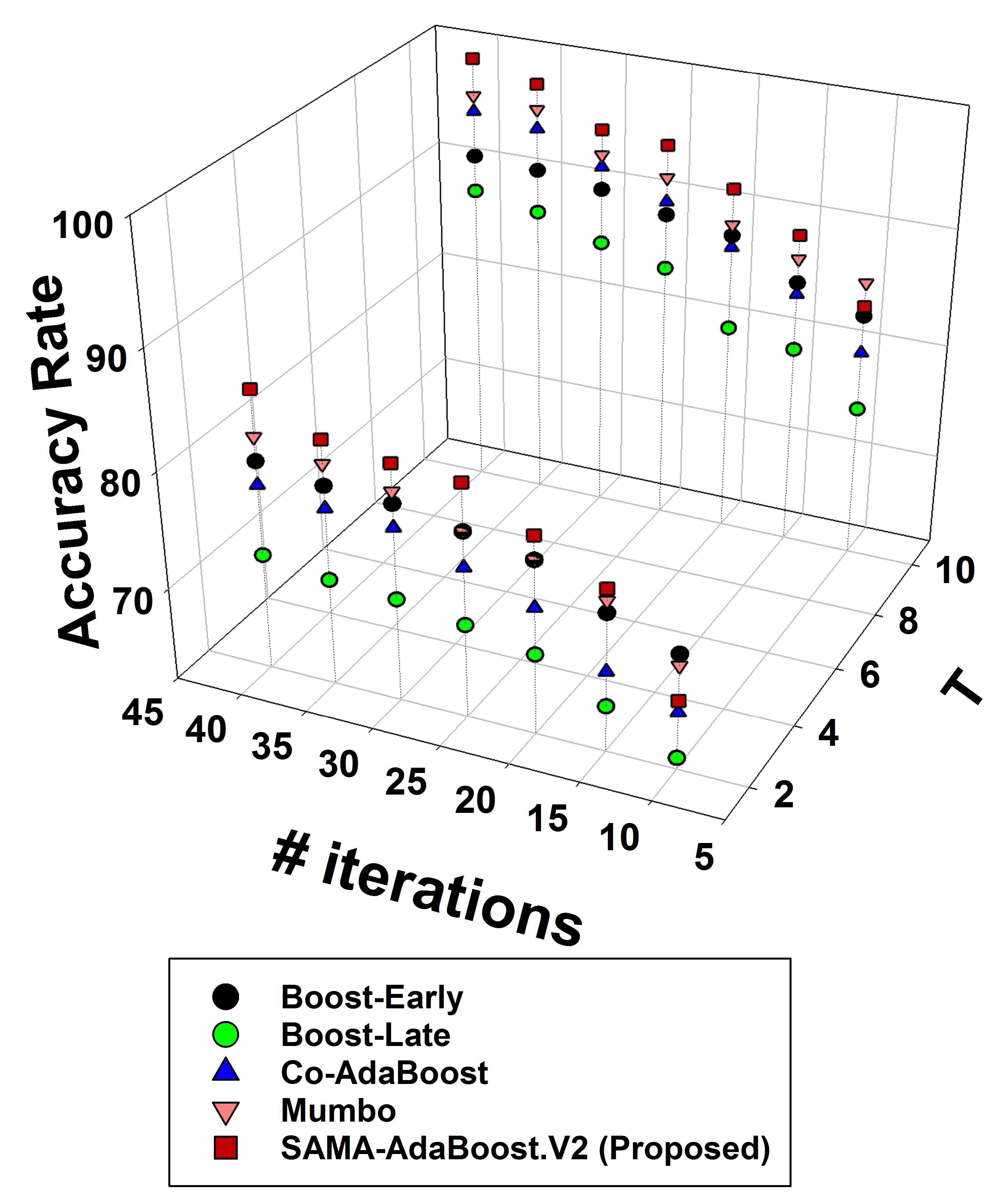}
\caption{Comparison of generalization accuracy rate of different ensemble classifiers for human eye classification. \textbf{T}: total boosting rounds. \textbf{\# iterations}: number of back propagation trainings per boosting rounds. }
\label{fig_eye_convergence}
\end{figure}

Under this 2-view setting we can compare SAMA-AdaBoost with Co-AdaBoost\cite{coadaboost}, 2-Boost\cite{boost2}, AdaBoost.Group\cite{adaboost.group} which support only 2-class, 2-view problems. 
For simulation purpose we use a 2-layer ANN with 5 units in hidden layer. Keeping less hidden nodes makes our baseline hypothesis `weak'. In Fig. \ref{fig_eye_convergence} we compare the classification accuracy rates of different boosted classifiers. 

Boost-Early refers to boosting on the entire \textit{144-D} feature space by concatenating features of SVD-Haar and SVD-HSV spaces. Boost-Late refers to separately boosting on individual feature space and final decision by majority voting. We use a pruned decision tree as another baseline on SVD-Haar space for 2-Boost. Co-AdaBoost tends to outperform other 2-class multiview boosting algorithms and thereby we report Co-AdaBoost's performance in Fig.\ref{fig_eye_convergence}. We see that at a fixed value of $T$, the rate of enhancement of accuracy rate with increase in number of ANN training iterations  is significantly higher for SAMA-AdaBoost compared to the competing algorithms. On average over ten rounds of boosting at 40 training iterations per round, accuracy rate of SAMA-AdaBoost is higher than that of MA-AdaBoost, Mumbo, Co-AdaBoost, 2-Boost, AdaBoost.Group, Boost-Early and Boost-Late by 2.3\%, 5.1\%, 6.2\%, 6.4\%,  6.9\%, 4\% and 10.1\% respectively. A ROC curve is a plot of true positive rate (TPR) at a given false positive rate (FPR). It is desirable that an ensemble classifier manifests a high TPR at a low FPR. For a good classifier the area under ROC curve (AUC), is  close to unity. 
\begin{table}
\caption{Comparison of area under ROC curve $\mathbf{(AUC)}$ and \textit{F-Score} $\mathbf{(F)}$ of different boosted classifiers for eye classification task after various rounds of boosting $\mathbf{(T)}$. In each boosting round the baseline ANNs have been trained for 40 iterations. Proposed SAMA-AdaBoost yields higher $AUC$ and $F$ compared to competing ensemble classifiers and thereby creating an ensemble space with better generalization capability.}
\centering
\begin{tabular}{l|*{3}{c}}\hline
\backslashbox{Algorithms}{Metrics}
&\makebox[2em]{T}&\makebox[3em]{$AUC$}&\makebox[3em]{$F$}
\\\hline\hline\\
SAMME \cite{samme}& &0.87 & 0.83\\
Boost-Late &&0.82&0.77\\
Boost-Early&&0.85&0.81\\
WNS \cite{wns}&&0.83&0.78\\
Zhang et al. \cite{prob_adaboost}&&0.86&0.82\\
Co-AdaBoost \cite{coadaboost}&5&0.86&0.84\\
2-Boost \cite{boost2}&&0.86&0.81\\
AdaBoost.Group \cite{adaboost.group}&&0.83&0.81\\
Mumbo \cite{mumbo1}&&0.88&0.87\\
MA-AdaBoost \cite{accv}&&0.90&0.88\\
\textbf{SAMA-AdaBoost (Proposed)}&&\textbf{0.93}&\textbf{0.91}\\\\

SAMME & &0.91 & 0.92\\
Boost-Late&&0.88&0.85\\
Boost-Early&&0.92&0.88\\
WNS&&0.90&0.87\\
Zhang et al.&&0.93&0.89\\
Co-AdaBoost&20&0.92&0.90\\
2-Boost&&0.90&0.89\\
AdaBoost.Group&&0.92&0.91\\
Mumbo&&0.93&0.92\\
MA-AdaBoost&&0.96&0.95\\
\textbf{SAMA-AdaBoost (Proposed)}&&\textbf{0.98}&\textbf{0.97}\\\\
\hline
\end{tabular}
\label{table_eye}
\end{table}
F-Score, $F$, is given by,
\begin{equation}
F=2\frac{precision\times recall}{precision+recall}
\end{equation}
A high precision requirement mandates that we compromise on recall and vice versa and thus alone precision or recall is not apt for quantifying performance of a classifier. F-Score mitigates this difficulty by calculating the harmonic mean of precision and recall. It is desirable to obtain a high F-Score from a classifier. From Table \ref{table_eye} we see that at a given round of boosting, $AUC$ and $F$ of SAMA-AdaBoost is higher compared to other competing algorithms. Table \ref{table_eye} bolsters our claim that ensemble space created by SAMA-AdaBoost fosters faster rate of convergence of generalization error rates compared to its competing counterparts.

\par Finally, in Table \ref{table_alexsvm}, we compare the performance of SAMA-AdaBoost with state-of-the-art techniques of other paradigm such as AlexNet \cite{alexnet} \footnote[1]{Available at \href{http://caffe.berkeleyvision.org/model\_zoo.html}{http://caffe.berkeleyvision.org/model\_zoo.html}}, which is a popular CNN architecture and SVM-2K \cite{svm2k}\footnote[2]{Available at \href{http://www.davidroihardoon.com/code.html}{http://www.davidroihardoon.com/code.html}}, which a state-of-the-art SVM algorithm for training on two views of dataset. Alexnet was trained for 50 epochs (error saturated after this) with batch size of 100 with stochastic gradient descent optimization. SAMA-AdaBoost was trained for 10 boosting rounds with 40  epochs per round. We see that SAMA-AdaBoost outperforms SVM-2K and manifests comparable results to AlexNet. But training time for SAMA-AdaBoost is only 13 minutes compared to 19 and 45 minutes for SVM-2K and Alexnet respectively.

\begin{table}
\caption{Compariosn of training time and classification accuracy rate on eye classification dataset.}
\centering
\begin{tabular}{l  c c}\hline\\
Algorithm & Training Time (mins) &  Accuracy Rate\\\hline\hline\\
\textbf{SAMA-AdaBoost}& 13 & 99.1\\
\textbf{(Proposed)}&&\\\\
AlexNet \cite{alexnet}&45&99.6\\\\
SVM-2K \cite{svm2k}&19&95.2\\
\hline
\label{table_alexsvm}
\end{tabular}
\label{table_uci_datasets}
\end{table}


\subsection{Simulation on UCI Datasets}
\subsubsection{\textbf{Comparison of Generalization Accuracy Rates}}
In this section we evaluate our proposed boosting algorithm on the benchmark UCI datasets which comprise of real world data pertaining to financial credit rating, medical diagnosis, game playing etc. The details of the eleven datasets chosen for simulation is shown in Table \ref{table_uci_datasets}. We randomly partition the homogeneous datasets into two subspaces for multiview algorithms and report the best results. We use a 2-layer ANN with 3 hidden units as baseline learner on each view. In each boosting round, ANNs are trained by back propagation 30 times. We cannot test multiclass datasets such as \textit{`Glass'}, \textit{`Connect-4}, \textit{`Car Evaluate'} and \textit{`Balance'} by \cite{coadaboost,boost2} and \cite{adaboost.group} because these algorithms only support 2-class problems. We also report the average training time per boosting round for each dataset using SAMA-AdaBoost using Matlab-2013 on Intel i-5 processor with 4 GB RAM @3.2 GHz. In Table \ref{table_uci_test_set} we report the generalization accuracy rates of different boosted classifiers after $T$=5, 10 and 20 rounds of boosting.

\begin{table}
\caption{UCI Datasets selected for simulation purpose. }
\centering
\begin{tabular}{l l c c }\hline
Dataset&\hspace{-3mm}\# of instances&\hspace{-3mm}\# of attributes&\hspace{-3mm}\# classes\\\hline\hline
Glass&214&10&7\\
Connect-4&67557&42&3\\
Car Evaluate&1728&6&4\\
Balance Scale&625&4&3\\
Breast Cancer&699&10&2\\
Bank Note&1372&5&2\\
Credit Approval&690&15&2\\
Heart Disease&303&75&2\\
Lung Cancer&32&56&2\\
SPECT Heart&267&22&2\\
Statlog Heart&270&13&2\\
\hline
\end{tabular}
\label{table_uci_datasets}
\end{table}

\begin{table*}
\begin{minipage}{\textwidth}
\begin{center}
\caption{Comparison of generalization accuracy rates on selected UCI datasets by different ensemble classifiers after various rounds ($T$) of boosting. In majority instances proposed SAMA-AdaBoost achieves higher accuracy rates compared to competing algorithms. We can compare \cite{coadaboost,boost2,adaboost.group} only on datasets involving 2-class classification.}
\begin{tabular}{l|*{12}{c}}\hline
\backslashbox{Algorithms}{Datasets~\footnote{Corresponding accuracy rates for SVM-2K\cite{svm2k} are ~95.1,~~~~ 94.8, ~~~~90.3,~~~ 89.9,~~~~97.1,~~~~93.4,~~~~96.1,~~~~94.3,~~~~~94.9,~~~~95.9~~~~~93.1}}
&\makebox[2em]{T}&\makebox[3em]{Glass}&\makebox[2em]{Connect-4}&\makebox[2em]{Car}&\makebox[2em]{Balance}&\makebox[1.5em]{Breast}&\makebox[2em]{Bank}&\makebox[2em]{Credit}&\makebox[2em]{Heart}&\makebox[2em]{Lung}&\makebox[2em]{SPECT}&\makebox[2em]{Statlog}
\\\hline\hline\\
SAMME \cite{samme}&&72.1&68.1&75.4&81.2&80.1&78.2&80.1&69.1&66.5&78.2&72.1\\
WNS \cite{wns}&&70.1&68.0&73.5&80.2&77.1&67.2&78.4&68.1&65.9&77.0&70.8\\
Boost-Late&&70.0&67.4&73.0&78.6&76.1&68.2&79.1&68.0&65.0&75.3&70.1\\
Zhang et al. \cite{prob_adaboost}&&73.2&70.4&75.3&80.9&82.1&80.7&80.1&72.3&69.8&80.0&75.4\\
Co-AdaBoost \cite{coadaboost}&5&-&-&-&-&83.1&81.0&81.2&72.0&68.1&81.1&76.0\\
2-Boost \cite{boost2}&&-&-&-&-&84.0&81.3&80.9&73.1&68.0&81.2&77.2\\
AdaBoost.Group \cite{adaboost.group}&&-&-&-&-&83.8&81.0&81.2&72.9&70.8&78.2&75.3\\
Mumbo \cite{mumbo1}&&74.3&75.4&74.3&78.2&84.3&75.1&83.2&74.3&75.4&81.2&78.1\\
MA-AdaBoost \cite{accv}&&75.1&76.2&75.9&80.8&85.2&82.1&84.1&77.2&78.9&83.1&80.9\\
\textbf{SAMA-AdaBoost (Proposed)}&&75.4&77.4&76.2&80.8&85.6&83.2&84.7&78.0&78.0&83.1&81.2\\\\

SAMME&&79.8&76.4&81.2&84.2&85.4&84.1&85.7&74.3&74.2&86.3&78.6\\
WNS&&    74.3&73.8&77.2&81.2&83.2&78.2&84.2&70.9&71.1&82.3&74.3\\
Boost-Late  &&73.3&70.9&77.0&80.6&82.9&76.2&83.2&72.1&71.9&82.9&75.1\\
Zhang et al.  &&73.2&70.4&75.3&80.9&85.4&86.4&87.9&77.6&76.9&87.0&81.9\\
Co-AdaBoost&10&-&-&-&-&83.9&83.4&86.1&73.9&73.9&85.4&78.9\\
2-Boost  &&-&-&-&-&87.0&87.3&86.7&75.4&73.6&86.1&78.7\\
AdaBoost.Group &&-&-&-&-&86.5&84.3&86.9&75.4&74.9&87.5&79.8\\
Mumbo&&81.6&83.2&78.3&80.2&89.3&80.1&89.2&81.7&82.3&85.2&81.9\\
MA-AdaBoost&&86.5&85.9&82.3&86.7&89.2&86.1&89.8&84.2&86.9&88.3&85.9\\
\textbf{SAMA-AdaBoost (Proposed)}&&87.3&87.1&84.3&88.0&91.2&87.2&91.2&87.2&88.1&91.1&87.9\\\\

SAMME&&91.3&90.9&91.2&92.1&94.3&92.1&90.0&92.8&86.8&93.5&91.9\\
WNS&&    90.0&88.8&90.5&91.7&92.6&91.2&89.0&92.0&84.3&92.1&91.0\\
Boost-Late  &&90.0&88.0&89.3&91.8&92.1&91.0&87.8&90.0&83.2&90.7&89.9\\
Zhang et al.  &&93.2&92.1&92.9&93.5&95.4&94.2&91.0&93.2&89.3&94.3&92.9\\
Co-AdaBoost&20&-&-&-&-&93.7&92.0&89.8&92.0&85.1&92.1&90.8\\
2-Boost             &&-&-&-&-&93.0&92.8&90.0&93.5&88.1&94.1&92.1\\
AdaBoost.Group &&-&-&-&-&93.2&91.8&90.2&91.9&87.2&92.5&92.0\\
Mumbo&&95.2&94.0&89.2&90.2&98.0&90.2&95.4&94.1&95.8&96.9&95.0\\
MA-AdaBoost&&97.0&95.2&92.9&95.4&98.3&95.4&97.8&95.8&97.2&98.0&96.1\\
\textbf{SAMA-AdaBoost (Proposed)}&&98.3&97.3&94.3&96.5&99.1&95.2&99.0&97.4&98.1&99.2&97.9\\
\hline
\end{tabular}
\end{center}
\end{minipage}
\label{table_uci_test_set}
\end{table*}

\par We can see from Table \ref{table_uci_test_set} that our proposed SAMA-AdaBoost outperforms the competing boosted classifiers in majority instances. It is interesting to note that although Mumbo performs comparable to SAMA-AdaBoost on majority datasets, the performance of Mumbo degrades on  `Balance, `Car' and  `Bank'  datasets. These datasets are represented over a very low dimensional feature space. Disintegration of this low dimensional feature space into two sub spaces fails to provide Mumbo with a \textit{`Strong'} view. As mentioned before, success of Mumbo depends on the presence of a \textit{`Strong'} view which is aided by \textit{`Weak'} views. Co-AdaBoost and 2-Boost offers comparable performance on the datasets and tends to outperform SAMME in majority instances. Performance of WNS is slightly worse compared to SAMME because WNS boosts on a subset of entire sample space without any correction factor to compensate for the reduced cardinality of sample space. But, Zhang et al. incorporated the correction factor and the performance is usually superior compared to SAMME. 

\subsubsection{\textbf{Kappa-Error diversity analysis}} 
\begin{figure*}
\centering
\vspace{-5mm}\hspace{-25mm}\subfigure {\includegraphics[scale=.036]{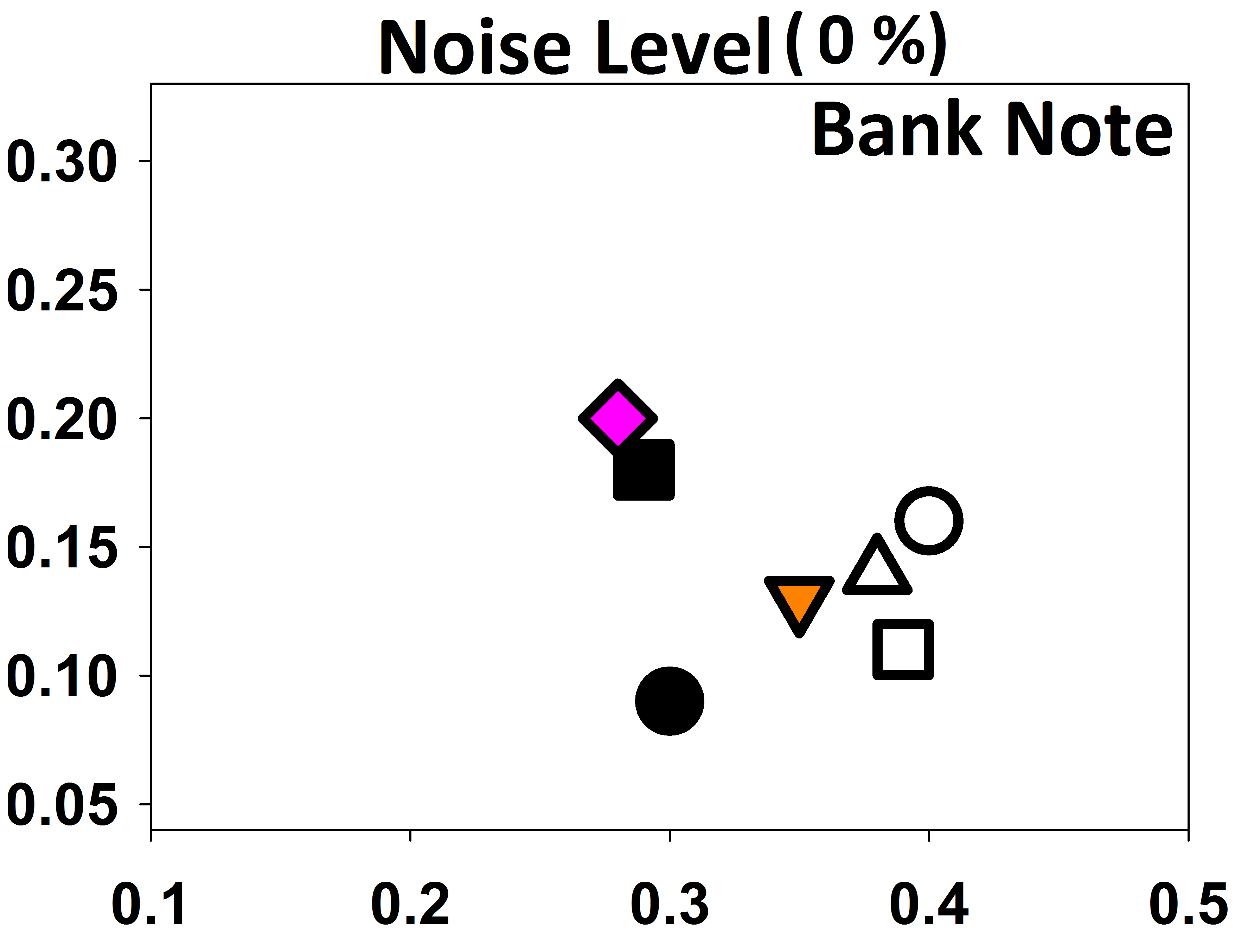}}
\subfigure {\includegraphics[scale=.036]{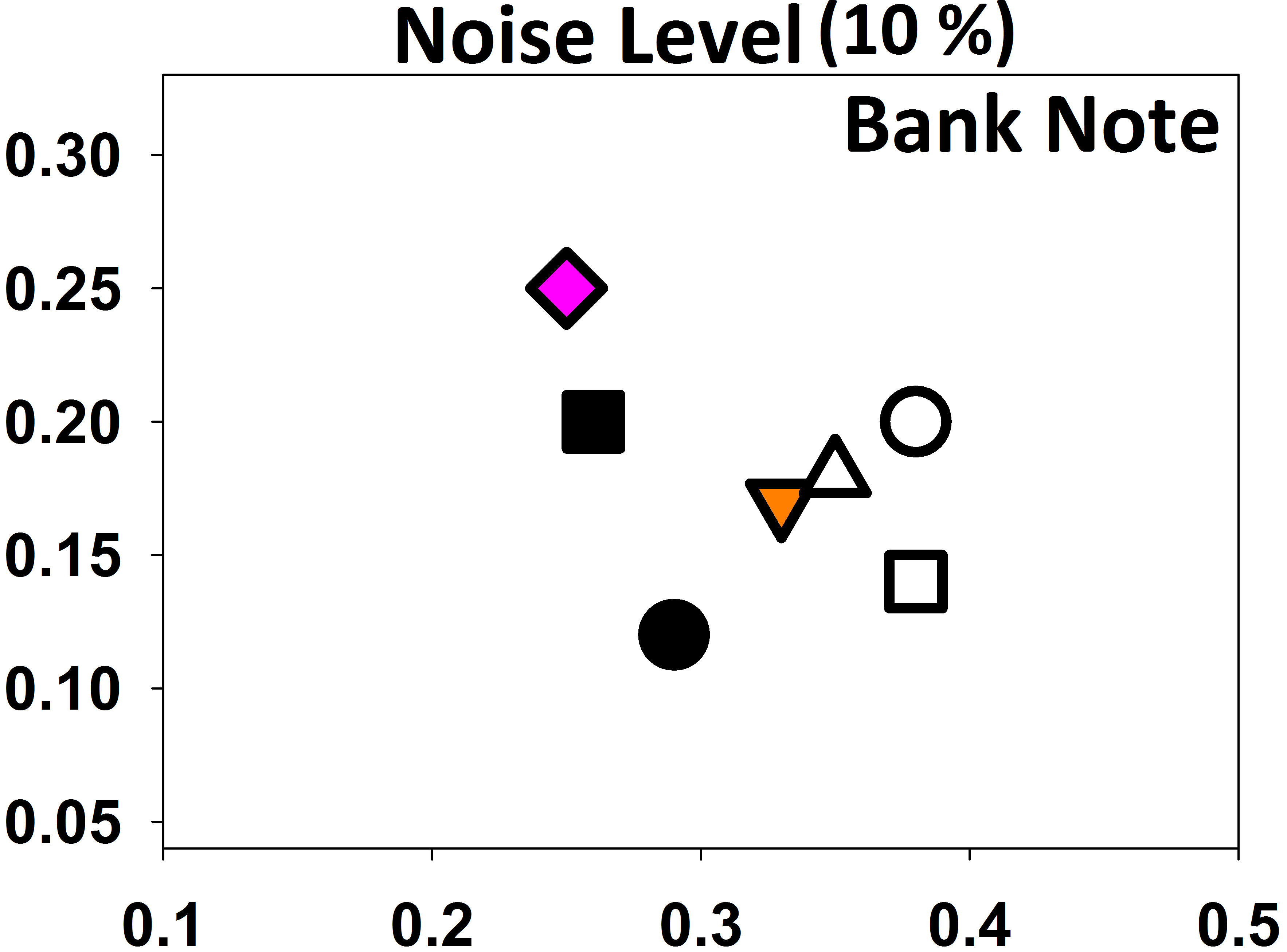}}
\subfigure {\includegraphics[scale=.036]{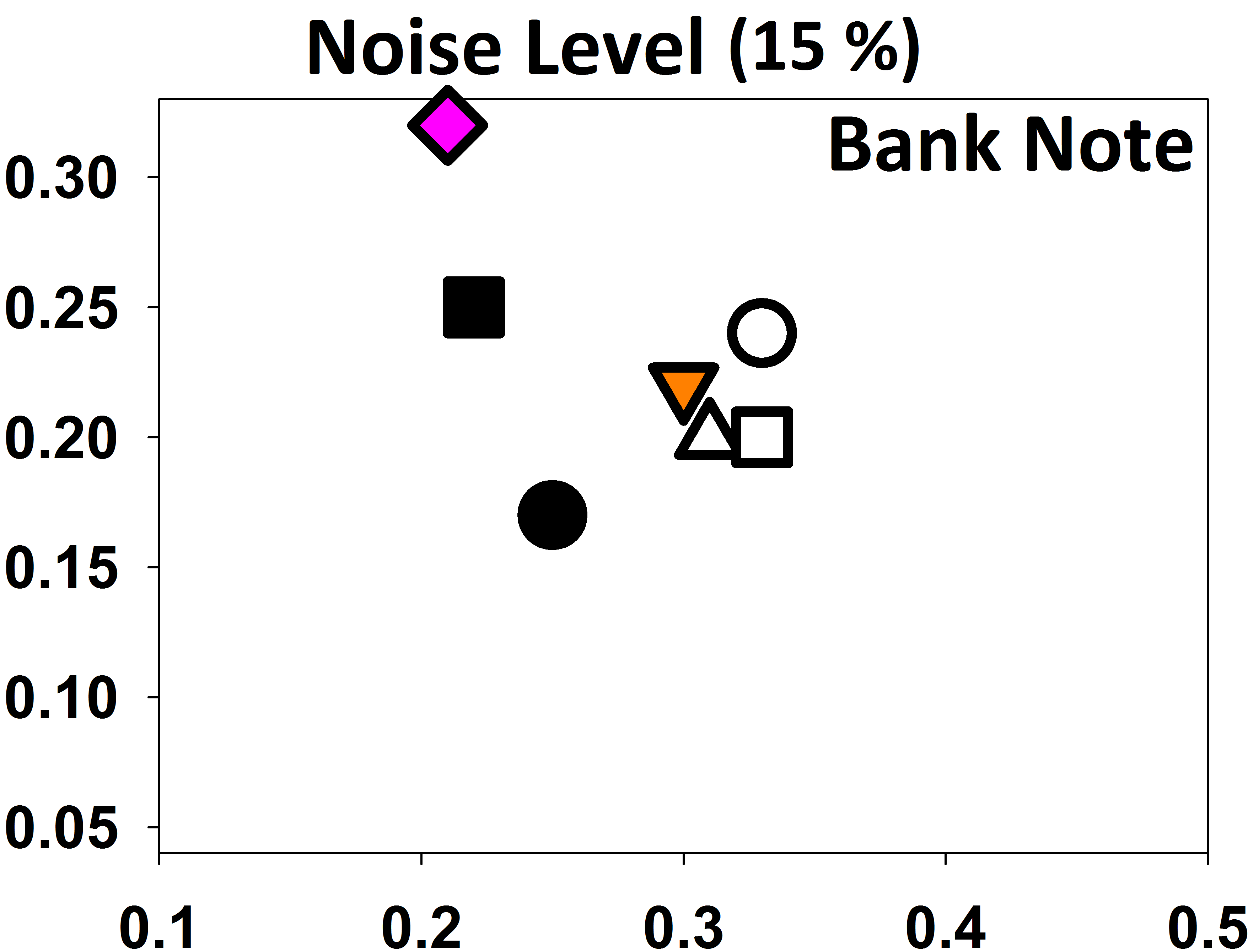}}\\
\hspace{-25mm}\subfigure {\includegraphics[scale=.036]{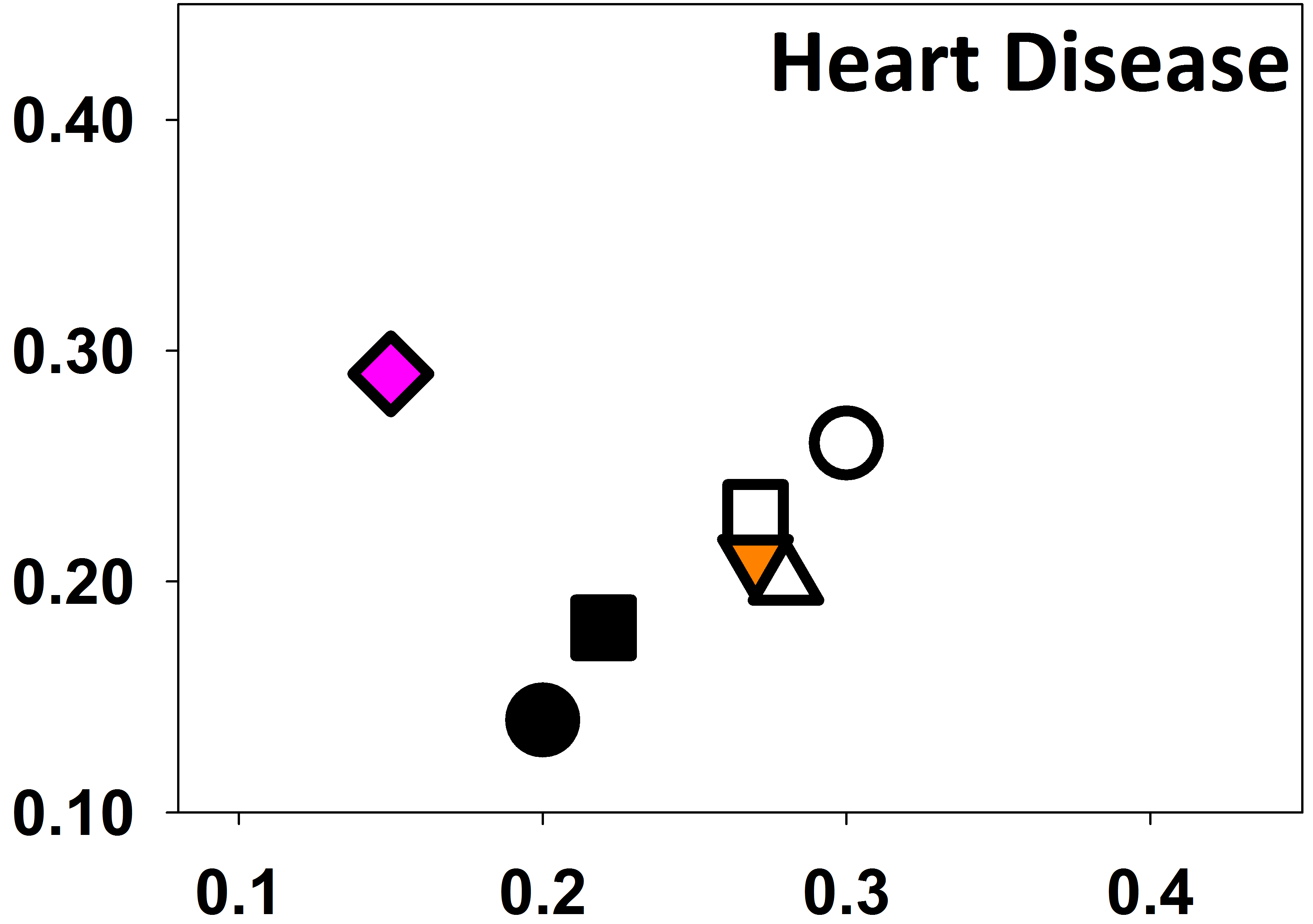}}
\subfigure {\includegraphics[scale=.036]{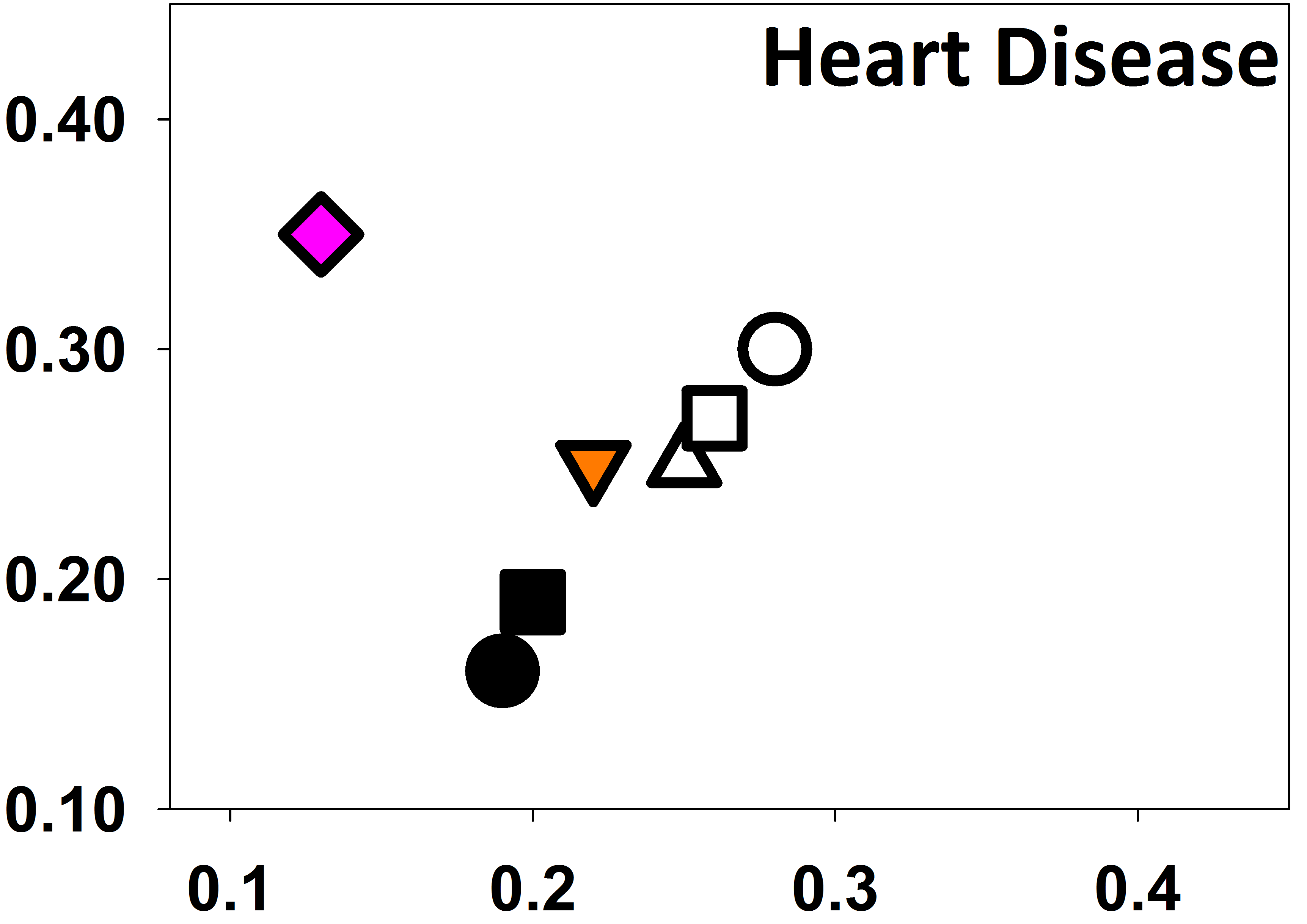}}
\subfigure {\includegraphics[scale=.036]{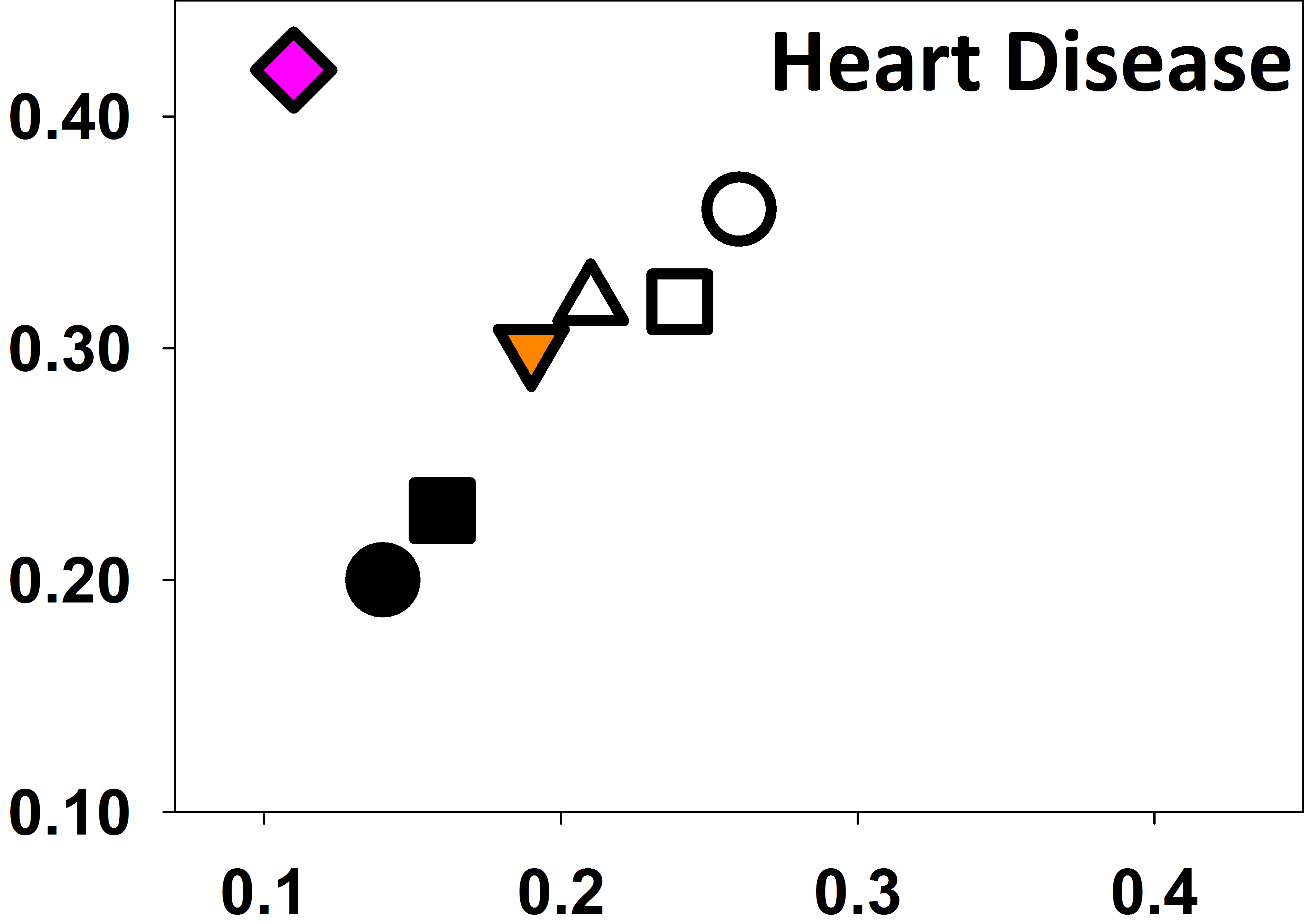}}\\
\subfigure {\includegraphics[scale=.036]{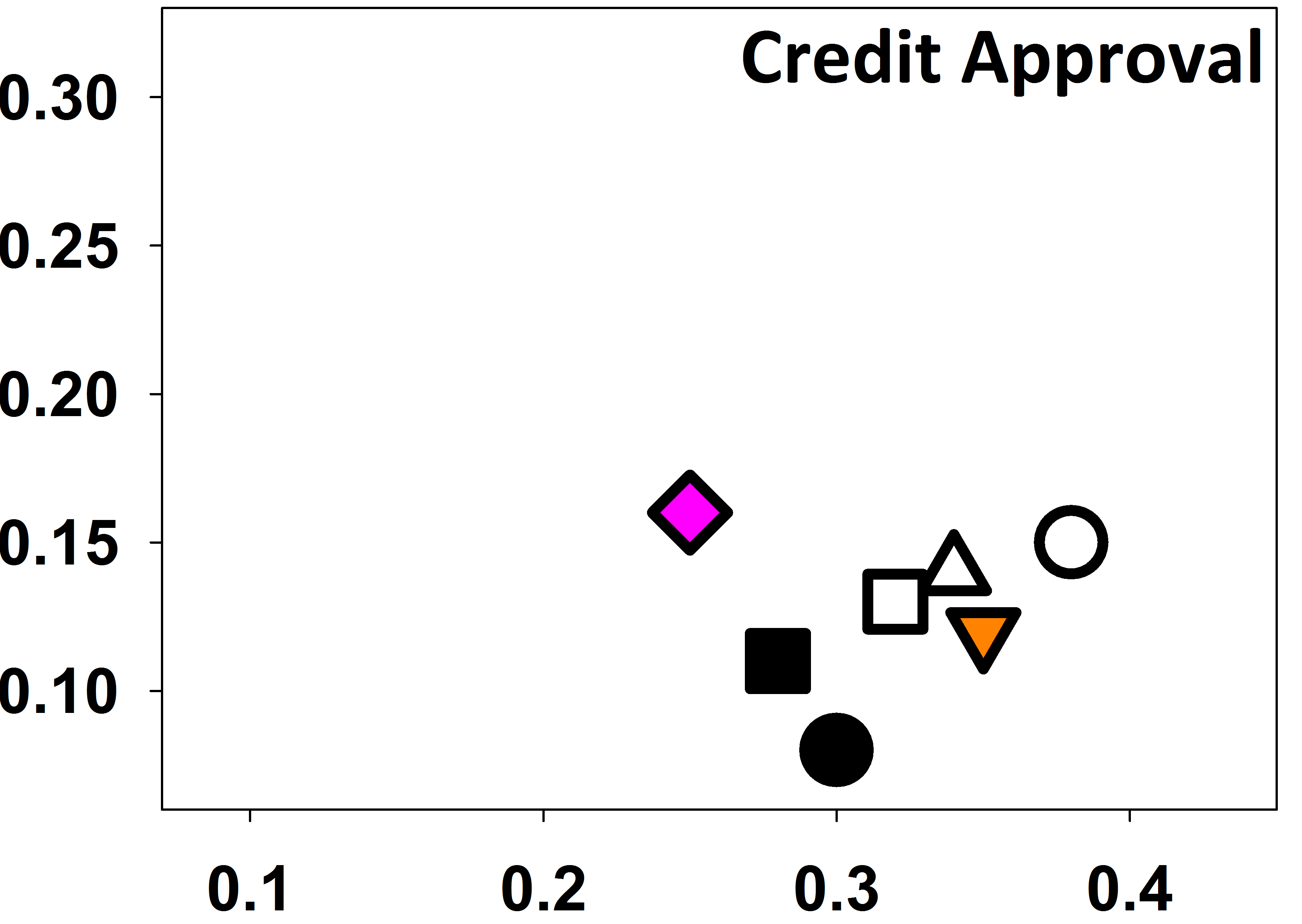}}
\subfigure {\includegraphics[scale=.036]{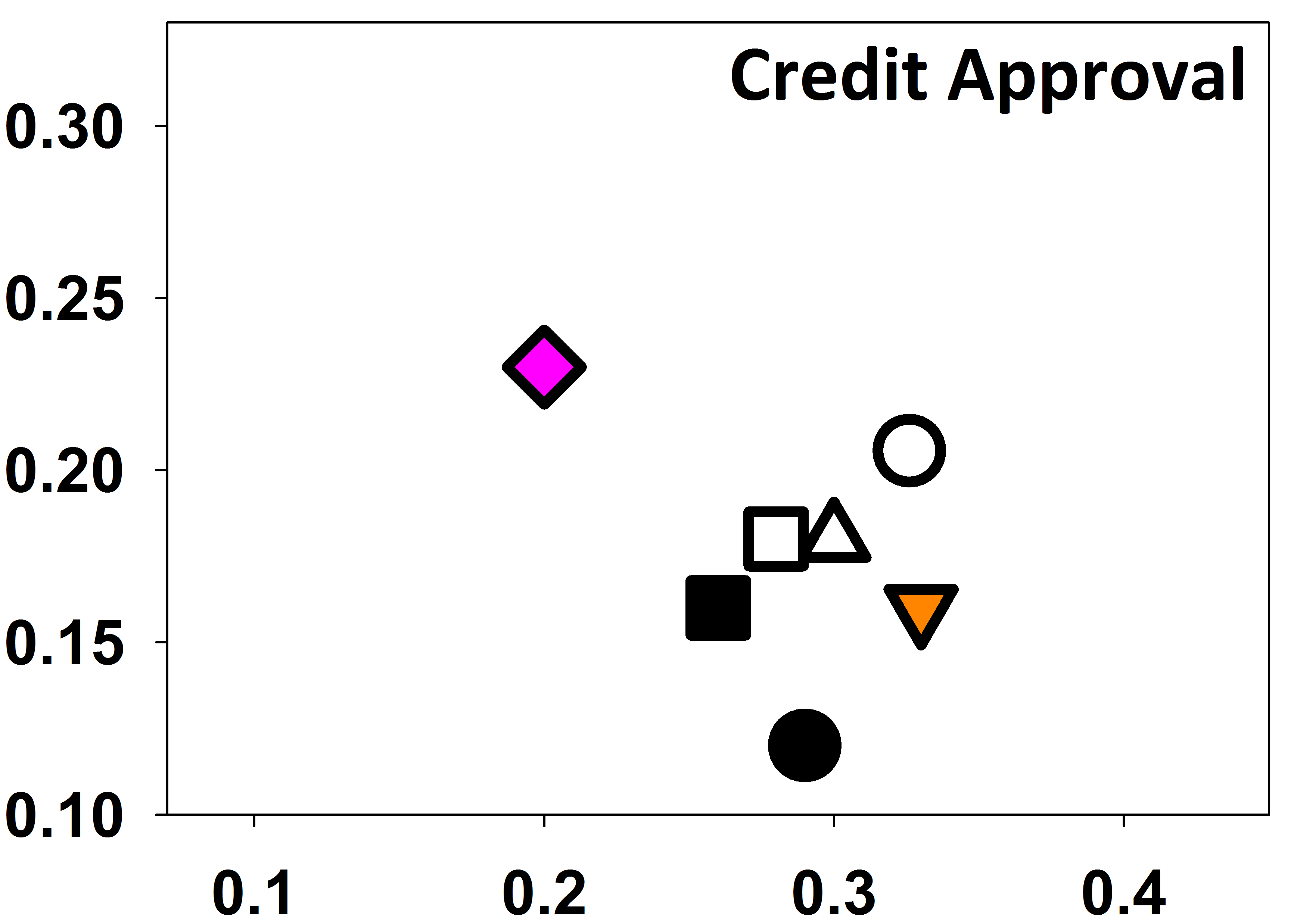}}
\subfigure {\includegraphics[scale=.036]{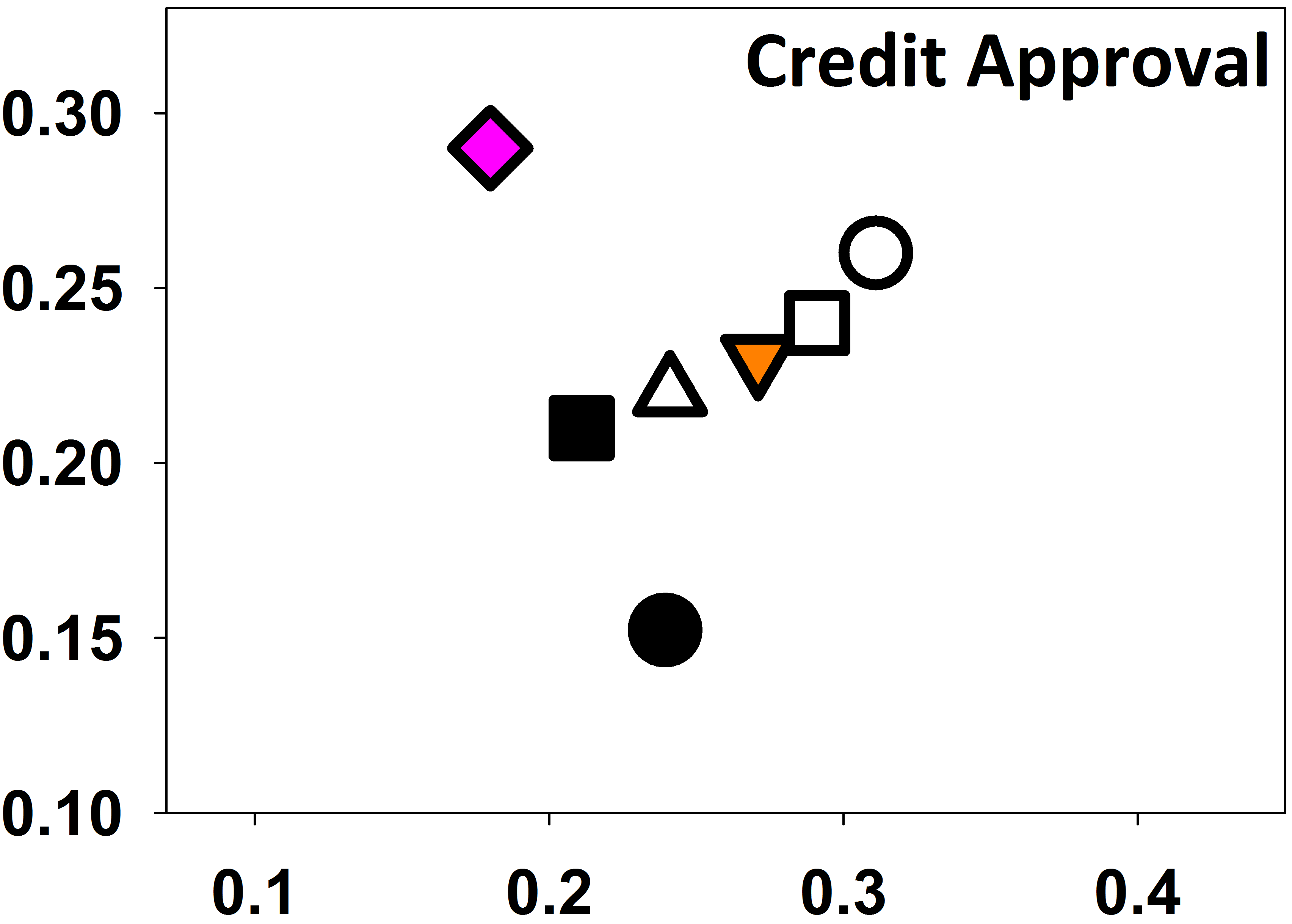}}
\subfigure {\includegraphics[scale=.25]{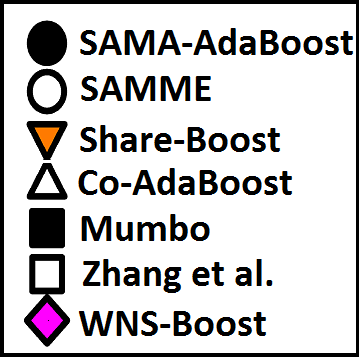}}\\
\caption{Kappa-Error diversity plots on UCI datasets for different boosted classifiers. For every possible pairwise combinations of member hypotheses $H_i$ and $H_j$ within an ensemble space we calculate the Kappa agreement coefficient $\kappa_{i,j}$ and  mean generalization error rate $E_{i,j}$ . \textbf{X axis:} centroid of $\kappa_{i,j}$. \textbf{Y axis:} centroid of $E_{i,j}$. Noise level indicates the fraction of original training labels that were perturbed before training the classifiers.}
\label{fig_kappa}
\end{figure*}

It is desirable that the individual members of an ideal ensemble classifier be highly accurate and at the same time the members should disagree with each other in majority instances \cite{kappa_support_2}. So, there is a trade-off between accuracy and diversity of an ensemble classifier space. Kappa-Error diagram \cite{kappa} is a visualization measure of error-diversity pattern of ensemble classifier space. For any two members $H_i$ and $H_j$ of ensemble space, $E_{i,j}$ represents average generalization error rates of $H_i$ and $H_j$ and $\kappa_{i,j}$ denotes the degree of agreement between $H_i$ and $H_j$. Define a coincidence matrix $M$ such that $M_{k,l}$ denotes the number of examples classified by $H_i$ and $H_j$ to classes $k$ and $l$ respectively. Kappa agreement coefficient $\kappa_{i,j}$ is then defined as,
\begin{equation}
\kappa_{i,j}=\frac{\frac{\sum_{p=1}^L\mathbf{M_{p,p}}}{m}-\sum_{l=1}^L\left [\sum_{m=1}^L\frac{\mathbf{M_{l,m}}}{m}\sum_{m=1}^L\frac{\mathbf{M_{m,l}}}{m}\right]}{1-\sum_{l=1}^L\left [\sum_{m=1}^L\frac{\mathbf{M_{l,m}}}{m}\sum_{m=1}^L\frac{\mathbf{M_{m,l}}}{m}\right]}
\end{equation}
where $L$ is the total number of classes. $\kappa_{i,j}=1$ signifies  $H_i$ and $H_j$ agree on all instances. $\kappa_{i,j}=0$ means  $H_i$ and $H_j$ agrees by chance while $\kappa_{i,j}\leq 0$ signifies agreement is less than expected by chance. Kappa-Error diagram is a scatter plot of $E_{i,j}$ v/s $\kappa_{i,j}$ for all pairwise combinations of $H_i$ and $H_j$. Ideally, the scatter cloud should be centered near lower left portion of the graph. Fig. \ref{fig_kappa} shows the Kappa-Error plots on three UCI datasets at different levels of labeling noise. We randomly perturb a certain fraction of training labels and train the classifiers on the artificially tampered datasets.
The plots are for classifiers trained over 15 rounds of boosting with 40 iterations of ANN training per round. So, we have total $^{15}C_2$ combinations of member learners. In Fig. \ref{fig_kappa} we plot only the centroids of scatter clouds of different classifiers because the scatter clouds are highly overlapping.
Fig. \ref{fig_kappa} reveals some interesting observations.

\textbf{1:} Scatter clouds of proposed SAMA-AdaBoost usually occupy the lowermost regions of the plots. This signifies that the average misclassification errors of members within SAMA-AdaBoost ensemble space is lower compared to competing ensemble spaces. Presence of such veracious members within SAMA-AdaBoost's ensemble space aids in enhanced classification prowess.
\textbf{2:} Scatter clouds of Mumbo on `Bank Note' dataset tends to be at a higher position compared to majority of other datasets. A relatively high position in Kappa-Error plot signifies an ensemble space consisting mainly of incorrect members. This observation also explains the degraded performance of Mumbo on `Bank Note' dataset as reported in Table \ref{table_uci_test_set}.
\textbf{3:} Addition of labeling noise shifts the error clouds to left and thereby enhancing diversity among the members. Simultaneously, the average error rates of the ensemble spaces also increase; this observation again highlights the error-diversity trade-off. 
\textbf{4:} Upward shift of the error clouds of SAMA-AdaBoost due to addition of labeling noise is relatively low compared to the error clouds of other ensemble spaces. Thus, SAMA-AdaBoost is more immune to labeling noise. 
\textbf{5:} WNS-Boost is most affected by labeling noise as indicated by its error clouds occupying top most position in the plots. 
\textbf{6:} Zhang et al. introduced a sampling correction factor to account for training boosted classifiers on a subset of original sample space. The correction factor aids them in achieving better generalization capability compared to SAMME and obviously much better compared to WNS-Boost which lacks such correction factor.

\subsection{Performance on MNIST dataset}
In this section we compare our algorithm on the well known MNIST hand written character recognition dataset which consists of 60,000 training and 10,000 test images. For multiscale feature extraction, we follow the procedures of \cite{mnist_tx}. Intially, images are resized to 28$\times$28. Next we extract three level hierarchy of Histogram of Oriented Features (HOG) with 50\% block overlap. The respective block sizes in each level are 4$\times$4, 7$\times$7 and 14$\times$14 respectively and the corresponding feature dimensions of the levels are 1564, 484 and 124. Features from each level serve as a separate view space for our algorithm. In Table \ref{table_mnist} we compare the performance of SAMA-AdaBoost with MA-AdaBoost, SAMME, Mumbo and Early-Boost. We use a single hidden layer neural network with $\frac{\sqrt{d}}{4}$ hidden nodes; where $d$ is feature dimensionality. In each boosting round, a network is trained for 30 epochs.

\begin{table}
	\caption{Test Set Error Rates on MNIST dataset. \textbf{T:} number of boosting rounds.}
	\centering
	\begin{tabular}{c c c c c c}\hline\\
		T& \hspace{-2mm}\textbf{SAMA-AdaBoost}\hspace{-2mm} &MA-AdaBoost\hspace{-2mm}&SAMME\hspace{-3mm}&Mumbo\hspace{-5mm}&Boost-Early\\
		&\textbf{(Proposed)}&\cite{accv}&\cite{samme}&\cite{mumbo_speech}&\\\hline\hline\\
		5&2.03& 2.20 & 2.38 & 2.35 & 2.41\\
	   10&1.10& 1.21 & 1.49 & 1.39 & 1.52\\
	   20& ~~0.80 \tablefootnote{We achieve an error rate of 0.7 using AlexNet \cite{alexnet} after 200 epochs}& 0.88 & 1.10 & 1.02 & 1.17\\
		\hline
	\end{tabular}
	\label{table_mnist}
\end{table}
It can be seen that the proposed SAMA-AdaBoost fosters faster convergence on generalization error rate. The observation furthers bolsters our thesis that multiview collaborative boosting is a prudent paradigm of multi feature space learning.

\section{Computational Complexity}
In this section we present a brief analysis on computational complexity of proposed SAMA-AdaBoost with neural network as base learner. The analysis is based on the findings of \cite{neural_complexity}. In Table \ref{table_neural_complexity} we elucidate the network specific variables which are used for complexity analysis.
\begin{table}
\caption{Parameters of Neural Network}
\centering
\begin{tabular}{c l}\hline
Symbol&Representation\\\hline\hline
$n$& cardinality of training space\\
$L$& total number of layers\\
$l$& Layer number $l$\\
$N_l$&Total activation nodes of layer $l$\\
$N_L$& Number of nodes in output layer\\
$w$&Total number of weights\\
$J_m$&Dimensionality of residual Jacobian $(N_L \times n)$\\
$W^l_{j,k}$ &Weight connection between node $j$ (layer $l$) \\&with node $k$ (layer $l+1$)\\
$z^l_{j,i}$&Activation of node $j$ of layer $l$ for example $x_i$\\\\
$\Omega^l_j (\cdot)$&Activation function of node $j$ in layer $l$\\
$A^l$&Cost of calculating total $\Omega^l(\cdot $) activations\\
$\epsilon^l_{j,i}$&$\frac{\partial E_i}{\partial z^l_{j,i}}$\\
$D_l$&Cost of calculating total $\Omega^{l'}(\cdot $) derivatives\\
$\Psi ^l$&Weight matrix connecting layer $l$ with $l+1$\\&Total elements = $[N_l+1 \times N_{l+1}]$\\
\hline
\end{tabular}
\label{table_neural_complexity}
\end{table}
We identify the key steps in both feed forward and backward pass and analyze the complexity individually. Refer to \cite{neural_complexity} for detailed explanation.
\vspace{-3mm}\subsection{Feed Forward}
\textbf{Step 1:} Complexity of Feeding Inputs to a Node \\
Cumulative input to node $J$ of layer $l (2 \leq l \leq L)$ for $x_q$ is given by:
\begin{equation}
\gamma^l_{j,q}=\sum_{i=1}^{N_{l-1}+1\Psi_{ij}^l-1 z^{l-1}_{i,q}}; j \in \{1,2,...N_l \}
\label{eq_step1}
\end{equation}

\textbf{Step 2:} Non Linear Activation of Node\\
Node $j$ of layer $l (2 \leq l \leq L)$ for $x_q$ is activated as:
\begin{equation}
z^l_{j,q}=\Omega^l_j(\gamma^l_{j,q})
\label{eq_step2}
\end{equation}

\textbf{Step 3:} Output Error Evaluation\\
With $o^N_{k,q}$ as $k^{th}$ ground truth label for $x_q$, squared error loss is defined as:
\begin{equation}
E(W)=0.5 \sum_{q=1}^n \sum_{k=1}^{N_L} (z^N_{k,q}-o^N_{k,q})^2
\end{equation}

\subsection{Backward Pass}
\textbf{Step 4:} Node Sensitivity Evaluation\\
At ouput node, sensitivity is for $x_q$ is given by:
\begin{equation}
\epsilon^N_{k,q}=\frac{\partial E_q}{\partial z^N_{k,q}}= z^N_{k,q} - o^N_{k,q}; k \in \{1,2,... N_L \}
\label{eq_step4_1}
\end{equation}
Sensitivity is propagated at backward layer, $l (1 < l < L)$ by the following recurrence:
\begin{equation}
\epsilon^l_{j,q}=\frac{\partial E_q}{\partial z^l_{k,q}}=\sum_{k=1}^{N_{l+1}} W^l_{j,k}\delta^{l+1}_{k,q}; j \in \{ 1,2,..N_l\}
\label{eq_step4_2}
\end{equation}
\begin{equation}
\delta^{l+1}_{k,q}=\epsilon^{l+1}_{k,q} \Omega^{s+1}_k (\gamma^{s+1}_{k,q})
\label{eq_step4_3}
\end{equation}

\textbf{Process 5:} Computing gradient\\
Gradient for $W^l_{j,k}$ for $x_q$ is given by:
\begin{equation}
\Delta^l_{jk,q}=z^l_{j,q}\epsilon^{l-1}_{k,q}\Omega^{s+1'}_k(\gamma^{l+1}_{k,q})
\label{eq_step5}
\end{equation}
with $j \in \{1,2,.. (N_l+1) \}$ and $k \in \{1,2,...N_{l+1} \}$\\

\textbf{Step 6: }Update of Parameters with step size $\eta$
\begin{equation}
W^l_{j,k}:=W^l_{j,k}- \eta \Delta^l_{j,k} ~~\forall j,k
\end{equation}

\begin{table}
	\caption{Complexity of training a neural network per epoch}
	\centering
	\begin{tabular}{l l}\hline
		\textbf{Feed Forward Pass}&\\\hline\\
		Process 1: & $2d\sum_{l=1}^L (N_{l-1}+1)N_l=2nw$\\\hline
		Process 2: & $n\sum_{l=1}^LA^lN_l$\\\hline
		Process 3: & $nN_L$ for evaluating residual\\
		           & $2nN_L$ for evaluating error\\\hline\hline
		\textbf{Backpropagation}& \\\hline\\
                   & $nN_L$ for Eq.(\ref{eq_step4_1})\\
        Process 4: & $+n\sum_{l=1}^{N-1}N_{l+1}(D_{l+1}+1)$ for Eq.(\ref{eq_step4_3})\\
                   & $+2n\sum_{l=2}^{N-1}N_l N_{l+1}$ for Eq.(\ref{eq_step4_2})\\\hline
        Process 5: & $2n\sum_{l=1}^L(N_l+1)N_{l+1}=2nw$\\\hline
        Process 6: & $2w$ for batch training\\\hline
		
	\end{tabular}
	\label{table_neural_complexity2}
\end{table}
Table \ref{table_neural_complexity2} provides a brief analysis of computational complexity of each step of training a neural network on a single epoch. Specifically, if per epoch complexity of a base learning network is $\mathcal{O}{(f(.))}$, then overall complexity for $T$ boosting rounds with $e$ epochs is $\mathcal{O}(Tef(.))$. Through parallel distributive learning, SAMA-AdaBoost can be trained on computationally cheaper networks compared to traditional boosting methods which agglomerate features from all views. To appreciate this fact, we focus on Process 2 of Table \ref{table_neural_complexity2}. Complexity of this step depends on input feature dimensionality, $N_1$ and number of nodes in hidden layer, $N_2$. SAMA-AdaBoost trains on separate feature spaces and thus feature dimensionality is scaled down appreciably. Also, we know that higher input dimensionality explicitly demands more hidden layer nodes for better feature representation. A rule of thumb is $N_2 \propto \sqrt{N_1}$. Thus distributed boosting inevitably reduces computational cost during feed forward pass. Process 4 and 5 also reveal that during back propagation of error derivatives, cardinality of connections between input and hidden layer and hidden and output layer plays a pivotal role in overall complexity. Thus we can conclude that a light weight network (less number of connections) has lower computational complexity and such networks can be trained parallely by multiview boosting such as SAMA-AdaBoost. Our algorithm has an extra step of optimizing an univariate convex loss function, Eq.(\ref{eq_to_optimize}) at end of an epoch. Computational complexity of this optimization is negligible compared to the complexities of feed forward and back propagation passes. For better appreciation of our claim we compare  per epoch training times in Table \ref{table_runtime}. The network architectures are exactly same as discussed previously in the respective dataset's section. We see that proposed SAMA-AdaBoost is considerably faster than SAMME which is the current state-of-the-art multiclass boosting algorithm. The gain in training time is vividly manifested on larger dataset such as MNIST. Thus SAMA-AdaBoost is more suited for large scale classification. Also, we notice that SAMA-AdaBoost is faster than contemporary state-of-the-art multiview boosting framework such as Mumbo. MA-AdaBoost manifests comparable runtime to SAMA-AdaBoost and is thus not compared.
\begin{table}
	\caption{Per Epoch Training Time (secs) of Competing Ensemble Classifiers on Challenging Datasets}
	\centering
	\begin{tabular}{l c c c}\hline
	Algorithm & 100 Leaves & Eye Classification & MNIST \\\hline\\
	SAMME \cite{samme}& 2.6 & 4.4 & 34.3\\
	WNS-Boost \cite{wns}& 2.4 & 3.3 & 32.0\\
	Mumbo \cite{mumbo_speech}& 1.8 & 2.7 & 18.3\\
	\textbf{SAMA-AdaBoost} & 1.2 & 2.0 & 13.7 \\
	\textbf{(Proposed)} & & & \\\hline
	\end{tabular}
	\label{table_runtime}
\end{table}

\section{Discussion and Conclusion}
The current work fosters three natural extensions.
\begin{itemize}
\item The weight update algorithm proposed in this paper depends on the classification performance of the overall ensemble space. An immediate extension will be to formulate a framework for weight update over each view space capturing both the local performance on that view and also the global performance. We believe such a weight update framework holds the key for further enhancing the learning rate 
\item Easy examples will be repeatedly correctly classified by majority views on every round of boosting. It will be an interesting attempt to identify the easy examples and remove those from the training sample space in future boosting rounds. Such an approach is envisioned to speed up the learning rate even further.
\item Outliers are usually "tough" examples which tends to be misclassified by majority views over the rounds of boosting. So, the proposed method can be used as a generic outlier detector for any classification or regression task.
\end{itemize}
\par Prior works on statistical viewpoints on boosting suggested that AdaBoost can be modeled by forward stagewise modeling to approximate 2-class Bayes rule and multi class Bayes rule \cite{samme}. Using a similar justification, we proposed a mathematical framework for multiview assisted boosting algorithm using a novel exponential loss function. To the best of our knowledge, this is the first attempt to model a scalable boosting framework for multiview assisted multiclass classification using stagewise additive model.  The proposed model focuses on grading the difficulty of a training example instead of  imposing a simple `1/0' loss on a weak learner. Such a grading policy aids an ensemble space to concentrate more on `tougher' misclassified examples compared to `easier' misclassified examples. Our previous work \cite{accv} was primarily based on intuitive concepts and lacked a rigorous mathematical treatment. The proposed SAMA-AdaBoost converges at near optimum value of learning rate $\beta_t$ compared to \cite{accv} and thus SAMA-AdaBoost offers faster convergence rate on training set error and simultaneously achieves better generalization accuracy. We also provided analytical and numerical evidences to show that ensemble space of SAMA-AdaBoost has lower upper bound of empirical loss and higher confidence margin compared to ensemble space of MA-AdaBoost. Extensive simulations on plethora of datasets reveal the viability of the proposed model. The kappa-error analysis  demonstrates the robustness of our model to labeling noise. We would like to emphasize that though our proposed SAMA-AdaBoost demonstrates enhanced performance compared to traditional boosting and variants of multiview boosting, we are conservative to claim that SAMA-AdaBoost is the only viable viewpoint of multiview assisted multiclass boosting. Understanding the mechanism of boosting is still an open problem and interested researchers will be benefited to refer to \cite{discuss_boost}, in which the authors describe the short comings of additive model to describe boosting. However, considering the simplicity of implementation of SAMA-AdaBoost and its close resemblance to AdaBoost, we feel that SAMA-AdaBoost is a viable solution of boosting effectively on multiple feature spaces to create a superior ensemble classifier space compared to existing multiview boosting methods.

\bibliographystyle{IEEEtran}
\bibliography{IEEEabrv,ref_short}
\end{document}